\documentclass[letterpaper,twocolumn,10pt]{article}
\usepackage{usenix-2020-09}

\usepackage{tikz}

\usepackage{pifont}
\usepackage{subfigure}
\usepackage{caption}
\usepackage{subcaption}
\usepackage{algorithmic}
\usepackage{algorithm}
\usepackage{multirow}
\usepackage{graphicx} 
\usepackage{amsmath}
\usepackage{amsthm}
\usepackage{amssymb}
\usepackage{soul}
\usepackage{enumitem}
\usepackage{xcolor}
\usepackage{booktabs}
\newtheorem{definition}{Definition}[section]

\newcommand \ignore[1]{}
\newcommand{\newtext}[1]{\textcolor{black}{#1}}

\newcommand{\revisedadd}[1]{\textcolor{black}{#1}}
\newcommand{\reviseddelete}[1]{}

\newcommand{\reviseddeletetwo}[1]{}

\usepackage{filecontents}

\begin{document}

\date{}

\title{\Large \bf Disparate Privacy Vulnerability: Targeted Attribute Inference Attacks and Defenses}

\author{
{\rm Ehsanul Kabir}\\
Pennsylvania State University
\and
{\rm Lucas Craig}\\
Pennsylvania State University
\and
{\rm Shagufta Mehnaz}\\
Pennsylvania State University
} 


\maketitle

\begin{abstract}
As machine learning (ML) technologies become more prevalent in privacy-sensitive areas like healthcare and finance, eventually incorporating sensitive information in building data-driven algorithms, it is vital to scrutinize whether these data face any privacy leakage risks.
One potential threat arises from an adversary querying trained models using the public, non-sensitive attributes of entities in the training data to infer their private, sensitive attributes, a technique known as the attribute inference attack.
This attack is particularly deceptive because, while it may perform poorly in predicting sensitive attributes across the entire dataset, it excels at predicting the sensitive attributes of records from a few vulnerable groups, a phenomenon known as disparate vulnerability.
This paper illustrates that an adversary can take advantage of this disparity to carry out a series of new attacks, showcasing a threat level beyond previous imagination.
We first develop a novel inference attack called the disparity inference attack, which targets the identification of high-risk groups within the dataset.
We then introduce two targeted variations of the attribute inference attack that can identify and exploit a vulnerable subset of the training data, marking the first instances of targeted attacks in this category, achieving significantly higher accuracy than untargeted versions.
We are also the first to introduce a novel and effective disparity mitigation technique that simultaneously preserves model performance and prevents any risk of targeted attacks.
\end{abstract}

\section{Introduction}
Advancements in machine learning (ML) techniques have revolutionized the way data is analyzed and utilized, enabling the solution of complex problems and the development of a wide range of applications in many domains including privacy-sensitive ones, such as personalized healthcare~\cite{ahamed2018applying, kasula2023harnessing}, finance~\cite{culkin2017machine, dixon2020machine}, and customer analytics~\cite{artun2015predictive, chou2022predictive}. 
However, this technological leap has also launched a pivotal issue\textemdash the vulnerability of ML models to privacy attacks.
Recent studies reveal that ML models are vulnerable to various privacy breaches. For instance, the models may 
reveal whether specific data was used in their training process~\cite{shokri2017membership}, 
and even allow the deduction of confidential information from the training dataset~\cite{fredrikson2015model,mehnaz2022your, yeom2018privacy, carlini2021extracting}.
The second type of attack, namely, the model inversion attack, is particularly concerning as it enables adversaries to recover sensitive information from trained models.
This weakness has constrained the training and application of ML models in domains sensitive to privacy, where safeguarding the privacy of the data is paramount.

Model inversion attacks can be broadly categorized into two types: class representative reconstruction~\cite{zhang2020secret, yang2019neural} and attribute inference~\cite{fredrikson2015model, mehnaz2022your}. 
In class representative reconstruction, the adversary aims to construct representative data points for specific classes or categories of the training data.
In attribute inference attacks, notably suited for models utilizing tabular data, the attacker aims to identify specific attribute values within the training data.
It is especially disconcerting that tabular data, despite being the most widespread type of structured data, is far less investigated for vulnerabilities~\cite{fredrikson2015model, mehnaz2022your, jayaraman2022attribute, dibbo2023model} when compared with other kinds of data, e.g., images.
This data domain faces a profound privacy challenge from attribute inference attacks, where adversaries can ascertain sensitive attributes using the model’s predictions and the publicly accessible attributes of individuals whose data was leveraged for training.

\textbf{Motivation}.
Existing attribute inference attacks struggle to achieve high performance, introducing uncertainty into their predictions and thereby reducing the perceived severity of privacy leakage through ML models.
According to the evaluation by Jayaraman et al.~\cite{jayaraman2022attribute}, their performance is usually worse than that of an imputation attack, in which the attacker collects a small amount of auxiliary data (roughly 10\% the size of the target data) and applies data imputation techniques to estimate the missing sensitive attribute values.
This causes a further misjudgment of the threat level associated with attribute inference attacks.
However, the presence of disparate vulnerability across various groups within the dataset~\cite{mehnaz2022your, dibbo2023model} leads us to believe that the performance assessed on the whole dataset does not truly represent the amount of privacy leakage, as strong attack performance in some groups is offset by weak attack performance in others.
Consequently, if the adversary becomes successful in identifying the subsets of data with high attack performance, they can target those subsets and predict the sensitive attribute values with greater certainty.
Thus, we seek to answer the following research question: \emph{can an adversary effectively determine the vulnerability levels of different groups within the dataset and then perform targeted attacks that are highly effective?}

\textbf{Challenges}.
Identifying the dataset's most at-risk groups becomes a complex task under the assumption that the adversary lacks direct access to the training data and does not have a shadow dataset mirroring the training data's distribution, which is a realistic scenario in the context of model inversion attacks.
However, we show that the variation in factors causing disparity such as the correlation between the sensitive and output attributes among different groups of the dataset can be leveraged to identify the most vulnerable groups of the dataset.
To achieve this, the adversary needs to find a way to measure the variation in the factors causing disparity across different groups of the dataset.
To this end, we introduce a novel technique that leverages the confidence score distribution of the model's predictions on various groups of the dataset to assess the variation in factors causing disparity across these groups. 
Our approach introduces a metric called \emph{angular difference}, which can measure the vulnerability of a group.
We also discover that angular difference is closely related to the correlation between the sensitive attribute and the output at the group level and can be used to estimate this correlation.

\textbf{Proposed Attacks}.
We introduce a series of new attacks with the shared goal of exploiting the disparate vulnerability across groups. 
First, we propose an attack called \emph{disparity inference attack}, which, to the best of our knowledge, is the first of its kind. 
This attack aims to rank the groups of records according to their vulnerability to attribute inference attacks. 
This attack can assist an adversary in launching existing attacks by assigning a degree of certainty to predictions based on group membership.
\textcolor{black}{
Leveraging the disparity inference attack, we develop two novel targeted attribute inference attacks: single attribute-based and nested attribute-based, marking them as the first targeted attacks of their kind.
}
Through empirical evaluation, we show that these targeted attacks can attain substantially higher performance in terms of accuracy than their untargeted counterparts~\cite{mehnaz2022your} while necessitating far fewer queries to the target model.

The extensive and varied privacy leakage from exploiting disparate vulnerability brings forth the pressing question: \emph{what steps can be taken to mitigate disparity?}
Unfortunately, current defensive methods against attribute inference attacks are often ineffective or can even exacerbate disparity~\cite{dibbo2023model}.
To address this, first, we explore the existing mutual information regularization (MIR~\cite{wang2021improving}) defense and incorporate a disparity-aware objective into it, resulting in the Disparity-Aware Mutual Information Regularization (DAMIR) solution. We show that this disparity mitigation approach falls short in consistently achieving its goal. Consequently, we design a novel solution that mitigates disparity by balancing the contributing factors, which we term as \emph{Balanced Correlation Defense (BCorr)}. 
Our evaluation shows that BCorr consistently mitigates disparity while also maintaining the original task performance of the target model.

\textbf{Summary of contributions}.
Our work makes the following contributions:

  \ding{114} We present a novel attribute inference attack, termed the disparity inference attack, which aims to identify the most vulnerable groups in the training dataset. Our attack is the first to focus on identifying high-risk groups, and we show that our technique performs exceptionally well according to ranking similarity metrics.

  \ding{114} We are the first to explore targeted attacks within the attribute inference category, proposing two variations that focus on a vulnerable subset of the training data and achieve a significant performance boost over their untargeted counterparts.

  \ding{114} We introduce a novel disparity mitigation technique that effectively eliminates disparity between groups. At the same time, it preserves the target model’s performance and prevents targeted attribute inference attacks.

\section{Preliminaries}

\textbf{Attribute Inference Attack}.
Let $\mathbf{n}(x)$ denote the non-sensitive portion of a record $x$, and let $\mathcal{M}$ represent the target model. The objective of the attribute inference attack is to predict $s(x)$, the sensitive attribute value of $x$. Certain variations of the attack necessitate additional knowledge by the adversary, such as auxiliary data $D_{aux}$.

\noindent \textbf{Confidence Score-based Model Inversion Attack (CSMIA)}.
In this attribute inference technique introduced in \cite{mehnaz2022your}, an adversary aims to predict the sensitive value of record $x$ with class label $y$ by querying the model multiple times with $x_i$ where $n(x) = n(x_i)$ and $s(x_i) = s_i$, with $s_i$ representing the $i$-th sensitive value. The model returns predictions $y_i$ and confidence scores $conf_i$ for $i \in [1, k]$. If only one $y_i$ matches $y$, the corresponding $s_i$ is output. If multiple $y_i$ match $y$, the one with the highest $conf_i$ is chosen and the corresponding $s_i$ is output. Otherwise, the one with the lowest $conf_i$ is selected and its corresponding $s_i$ is output.

\noindent \textbf{Label Only Model Inversion Attack (LOMIA)}.
In the attribute inference technique introduced by \cite{mehnaz2022your}, the adversary generates predictions for $x_i$ similar to CSMIA but does not use confidence scores. They create an attack dataset from all $x$ that returned a true prediction for only one $x_i$, adding $(x, y)$ as input and $s_i$ as output. Subsequently, an attack model is trained and used to infer the sensitive attribute value on the remaining records.

\noindent \textbf{Imputation Attack}.
The adversary creates an attack dataset similar to LOMIA, but using $D_{aux}$. Subsequently, an attack model is trained to infer the sensitive attribute value on the records.

\noindent \textbf{Neuron Importance Attack}.
In this whitebox attack introduced in~\cite{jayaraman2022attribute}, $D_{aux}$ is utilized to identify the top 10 most correlated neurons within the MLP. For each record $x$, the weighted sum of the activation values of these top 10 neurons is calculated. If this sum exceeds a certain threshold, the attacker predicts that $x$ has the sensitive value of interest.

\noindent \textbf{Disparate Vulnerability of MIAI}.
Let,
$\mathcal{M}$ denote the MIAI attack model,
$\mathbb{D}$ denote the target dataset,
and $\mathcal{A}$ denote the attack algorithm that the adversary aims to launch.
Additionally let, $ASR(\mathcal{M}, \mathbb{D}, \mathcal{A})$ denote the attack success rate of launching $\mathcal{A}$ on model $\mathcal{M}$ and $\mathbb{D}$.
We state that $\mathcal{A}$ is \textit{disparate} if there exists two disjoint subsets $\mathbb{D}_1$ and $\mathbb{D}_2$ of $\mathbb{D}$ such that $|ASR(\mathcal{M}, \mathbb{D}_1, \mathcal{A}) - ASR(\mathcal{M}, \mathbb{D}_2, \mathcal{A})| > \epsilon$ for some $\epsilon > 0$. 
In other words, the attack success rate of $\mathcal{A}$ on $\mathbb{D}$ is not uniform across all subsets of $\mathbb{D}$.
Here, $\epsilon$ is a threshold below which any disparity is considered negligible.

\section{\revisedadd{Attack }Threat Model}
We assume the following  adversary capabilities:

    \ding{111} Access to the black-box target model, i.e., the adversary can query the model with $x$ and obtain the output label $y$ and the corresponding confidence scores.

    \ding{111} Full knowledge of the non-sensitive attributes. 

    \ding{111} Knowledge of every possible value of the sensitive attribute and any non-sensitive attributes the adversary regards as group attributes.

\newtext{
These capabilities are considered realistic in the context of model inversion attacks, with most current model inversion attacks assuming at least these capabilities.
Notably, the attacks introduced in Yeom et al~\cite{yeom2018privacy}, Fredrikson et al~\cite{fredrikson2015model} and CSMIA require full non-sensitive attribute knowledge for the specific target record x, whereas LOMIA needs complete non-sensitive attribute information for the entire target dataset.
}
Unlike in previous works~\cite{fredrikson2015model, yeom2018privacy}, the adversary in this case does not need to be aware of the marginal priors, defined as the relative frequencies of the sensitive attribute values, to conduct the attack.
In addition, the adversary can perform the attack without needing an auxiliary dataset. 
This differs from most existing attribute inference attacks~\cite{jayaraman2022attribute,yeom2018privacy}, which typically require an auxiliary dataset that matches the distribution of the target dataset, except for CSMIA and LOMIA~\cite{mehnaz2022your}.
\revisedadd{
The attacker is considered to have complete knowledge of all possible non-sensitive attribute values. This assumption is realistic because publicly queryable ML models often reveal all possible values of query attributes. Leveraging this information and our proposed targeted attribute inference attacks, the attacker can identify groups based on different non-sensitive attributes.
When performing attacks, we assume the adversary has fewer capabilities to investigate the extent of privacy leakage under practical constraints. 
Conversely, during the evaluation of our defense, we consider an adversary with greater capabilities to rigorously evaluate the defense’s strength, as outlined in the threat model in section~\ref{sec:bcorr}.
}

\section{Uncovering High-Risk Groups}
\label{sec:uncovering_vulnerability}
\subsection{Key Factor Contributing to Vulnerability}.
To identify groups with a high risk of privacy leakage, it is essential to understand the factors contributing to the differing vulnerability levels among records from high-risk groups and those from low-risk groups.
The core factor contributing to the vulnerability of ML models to attribute inference attacks is the association between input and output data. 
For a model to accurately predict outputs during inference, it must learn the associations present in the training data. 
Therefore, a strong association between input and output data in the training set increases the likelihood of model inversion attacks, where an adversary uses the model to infer sensitive attribute values in the training data.
One basic way to measure the association between two variables is through correlation, leading to the natural assumption that the correlation between the sensitive attribute and the output plays a crucial role in the vulnerability to attribute inference attacks.
\newtext{
By ‘correlation’, we mean Pearson’s correlation, which is often used interchangeably in this domain.
}
We conduct a simple experiment to provide evidence supporting this hypothesis.

\begin{figure*}
    \centering
    \begin{minipage}{0.23\textwidth}
         \includegraphics[width=\linewidth]{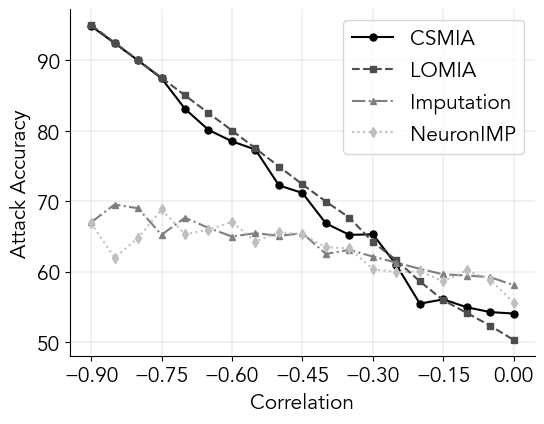}
          \caption*{(a)}
          \label{fig:corr_vs_acc_census}
    \end{minipage}
    \begin{minipage}{0.23\textwidth}
         \includegraphics[width=\linewidth]{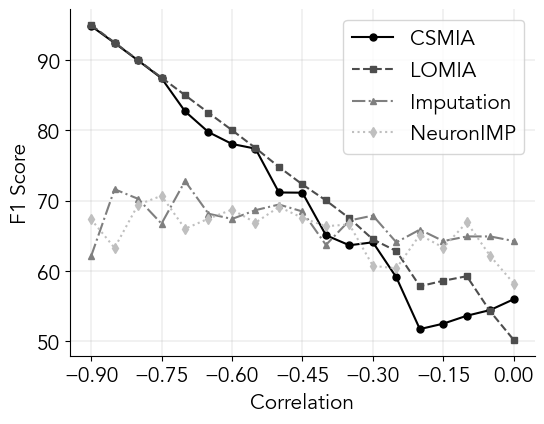}
          \caption*{(b)}
          \label{fig:corr_vs_f1_census}
    \end{minipage}
    \begin{minipage}{0.23\textwidth}
         \includegraphics[width=\linewidth]{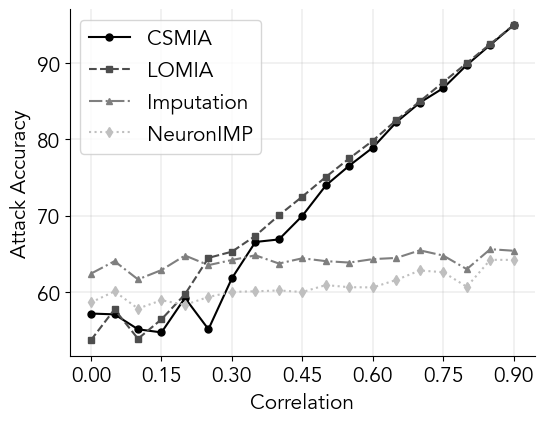}
          \caption*{(c)}
          \label{fig:corr_vs_acc_texas}
    \end{minipage}
    \begin{minipage}{0.23\textwidth}
         \includegraphics[width=\linewidth]{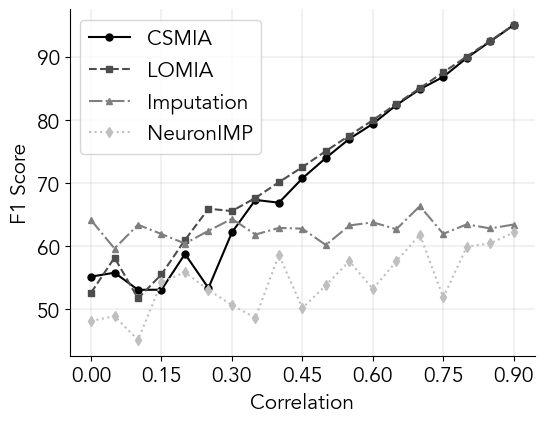}
          \caption*{(d)}
          \label{fig:corr_vs_f1_texas}
    \end{minipage}
\caption{Comparative evaluation of CSMIA, LOMIA, Imputation Attack, and NeuronImportance Attack across scenarios where dataset have varying level of correlation ranging from $0$ to $-0.9$ for Census19 (a-b) and $0.9$ for Texas-100X (c-d)}
\label{fig:corr_vs_attack_performance}
\end{figure*}

\textbf{Experiment Setup}.
We use the Census19 and Texas-100X datasets (detailed in section~\ref{sec:experimental_setup}) for this experiment and apply a sampling technique (described in section~\ref{sec:experimental_setup}) to create 19 training sets from each dataset, each with varying levels of correlation. 
We then train target models on each dataset and perform CSMIA~\cite{mehnaz2022your}, LOMIA~\cite{mehnaz2022your}, Imputation~\cite{jayaraman2022attribute}, and NeuronImportance~\cite{jayaraman2022attribute} attacks on these models.
\newtext{
For the latter two attacks, we assume the adversary utilizes a single auxiliary dataset across all scenarios, with a distribution identical to that of the original datasets.
}

\textbf{Results}.
Figure~\ref{fig:corr_vs_attack_performance} reports the accuracy and F1 scores of the attacks.
The plot offers several compelling observations.
In the Census-19 dataset, both CSMIA and LOMIA exhibit a monotonically increasing trend within the correlation range of -0.2 to -0.9, a pattern not observed in the Imputation and NeuronImportance attacks. 
A similar trend is seen in the Texas-100X dataset for correlation values ranging from 0.3 to 0.9.
The main factor behind the poor performance of Imputation and NeuronImportance at higher correlation magnitudes is that their auxiliary data does not have the same high correlation as the training data.
\newtext{
The performance of an imputation attack varies greatly depending on whether the auxiliary data shares the same distribution as the original data, an unrealistic scenario, or has a different distribution, which is more likely in practice. 
This distinction is discussed thoroughly in Section~\ref{sec:ideal_vs_practical_imputation}.
Interestingly, correlation’s effect is sign-agnostic; whether negative or positive, high correlations lead to increased vulnerability. 
This occurs because different labeling methods can alter the correlation sign.
An alternative labeling method (e.g., switching positive and negative outputs) can convert a previously negative correlation to a positive one.
The key takeaway from the results is that correlation is a significant factor in vulnerability to attribute inference attacks, prompting the question: does correlation also influence disparate vulnerabilities among groups? 
We explore this with a brief experiment.
}
Throughout the rest of this paper, we describe correlation as high/low to refer to its magnitude, avoiding redundancy and improving readability. 
Similarly, ‘correlation’ is used as shorthand to specifically refer to the correlation between the sensitive attribute and the output. 

\begin{figure*}[t]
    \centering
    \subfigure[CSMIA strategy]{
         \includegraphics[width=0.47\textwidth]{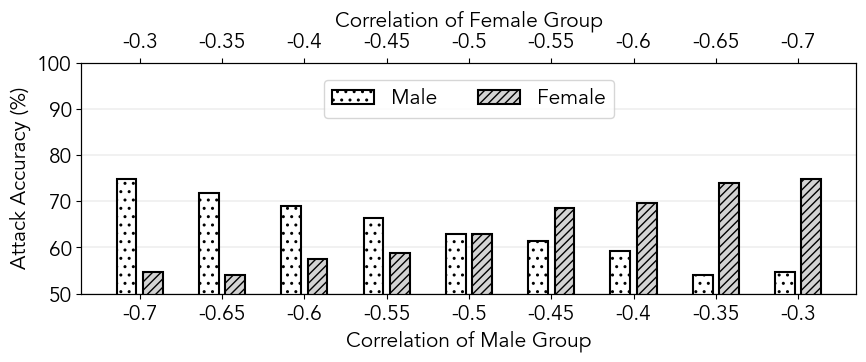}
         \label{fig:corr_vs_disparity_CSMIA}
    }
    \hfill
    \subfigure[LOMIA strategy]{
         \includegraphics[width=0.47\textwidth]{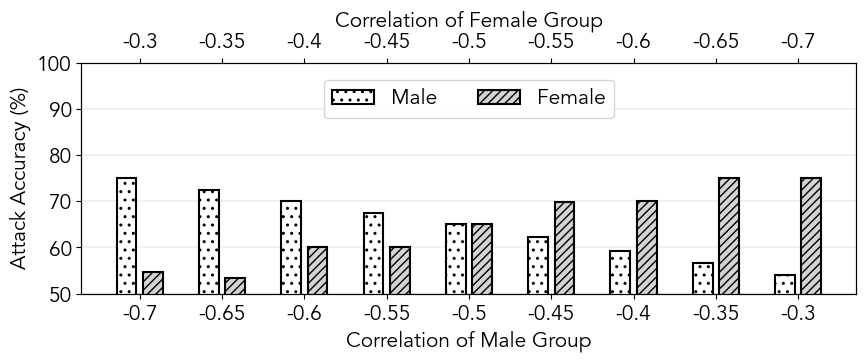}
         \label{fig:corr_vs_disparity_LOMIA}
    }
\caption{Correlation vs. Attack performance for Male and Female group for 9 different scenarios using Census-19 Dataset.}
\label{fig:corr_vs_disparity}
\end{figure*}

\noindent\textbf{Impact of Correlation on Disparate Vulnerability}.
We conduct another brief experiment to assess the impact of correlation on disparate vulnerability.
This experiment is carried out on the Census-19 dataset, using the \texttt{SEX} attribute to divide the dataset into Male and Female groups. 
We explore 9 scenarios, progressively increasing the correlation in Female records and decreasing it in Male records.
The experimental results are depicted in Figure~\ref{fig:corr_vs_disparity}.
The results clearly show that the impact of correlation on vulnerability is significant at a group level. 
When the correlation is high in Male records and low in Female records, the attack performance is high in the Male group and low in the Female group, and vice versa.
Both CSMIA and LOMIA performances exhibit this trend.
To emphasize, the experimental results demonstrate that correlation is a key factor influencing the varying vulnerabilities among groups within a dataset.

Given this observation, the key question is\textemdash \emph{Can an adversary having only black-box access to the model compare the correlation among these groups and thus identify the groups with high privacy risk?}
Note that precise measurement of correlation is not mandatory; being able to compare correlation between groups is sufficient.
However, comparing correlation among groups poses a challenge since the adversary in our threat model lacks access to an auxiliary dataset that matches the distribution of the training data.
To the best of our knowledge, no method exists to estimate correlation in the training data, let alone compare correlation among groups.

\begin{figure*}
    \centering
    \begin{minipage}{0.32\textwidth}
         \includegraphics[width=\linewidth]{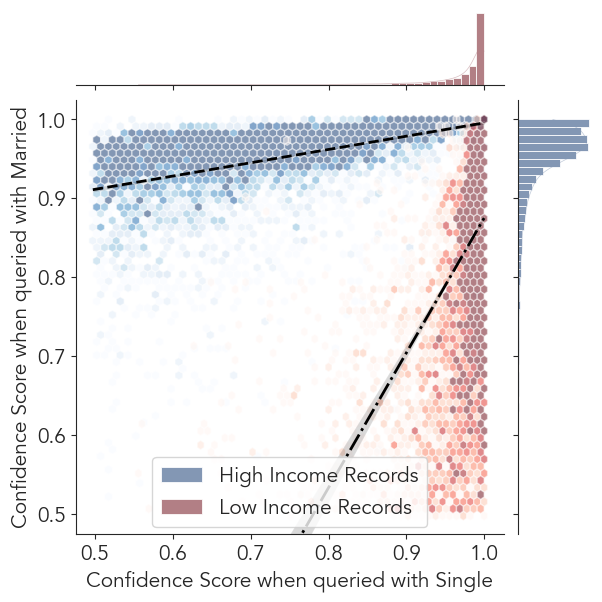}
          \caption*{(a) Group Correlation = -0.6}
          \label{fig:hexbin_-0.6}
    \end{minipage}
    \begin{minipage}{0.32\textwidth}
         \includegraphics[width=\linewidth]{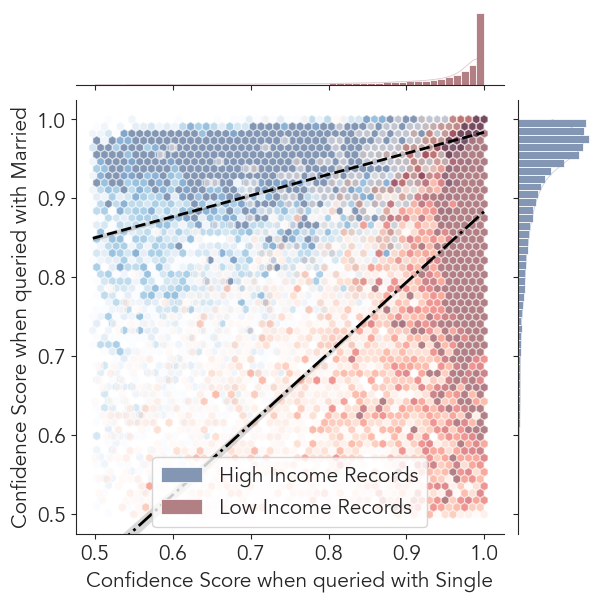}
          \caption*{(b) Group Correlation = -0.5}
          \label{fig:hexbin_-0.5}
    \end{minipage}
    \begin{minipage}{0.32\textwidth}
         \includegraphics[width=\linewidth]{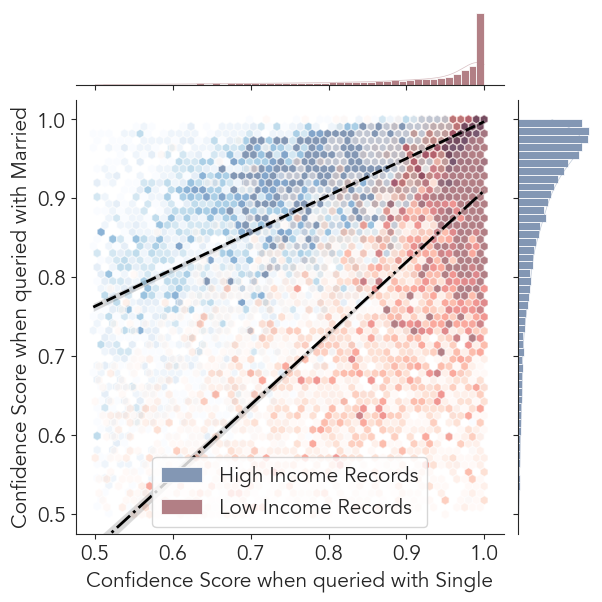}
          \caption*{(c) Group Correlation = -0.4}
          \label{fig:hexbin_-0.4}
    \end{minipage}
\caption{Histograms (bivariate and univariate) of confidence scores generated by querying records with different sensitive values, taken from groups with varying correlation levels in the Census19 dataset.}
\label{fig:hexbin}
\end{figure*}

\subsection{Comparing Correlation between Groups}
A high degree of correlation in the dataset indicates that certain values of the sensitive attribute, or certain ranges of values if the sensitive attribute is non-discrete, are more frequently associated with a specific class label compared to other values of the sensitive attribute. 
Conversely, a low correlation implies that the relative difference in the occurrence of sensitive attributes for a particular class label is minor or insignificant.
Our primary intuition is that during training, the target model is influenced by the relative frequency of sensitive attributes within a class label, leading it to predict with higher confidence for a record from that label when the sensitive attribute is set to the most frequent value, rather than to less common values.
In other words, the confidence score gap, defined as the difference in confidence scores generated by querying the same record with different sensitive attribute values, depends on the relative frequency of those values.
Since the level of correlation determines relative frequency, the distribution of the aforementioned confidence gap serves as an indicator of the correlation level, which could be used by an adversary. 
First, we will consider a dataset with a binary sensitive attribute and a binary output class to present our argument more clearly. 
Afterwards, we will discuss how the argument can be extended to multi-category or non-discrete sensitive attributes and multi-class outputs.
Let's assume the sensitive attribute values are `yes' and `no', and the output class values are `True' and `False'.
If the dataset has a strong correlation, the `True' class will contain significantly more `yes' records than `no' records.
As a result of this imbalance, the model will develop a bias towards assigning higher confidence to the `True' class for queries with the sensitive attribute set to `yes' compared to `no' for the same record.
If the records in a group exhibit a high degree of imbalance, the expected confidence gap will be large; conversely, a low degree of imbalance will result in a smaller expected confidence gap.
Thus, if an adversary analyzes the distribution of the confidence gap across multiple groups, they can predict which groups are more vulnerable.

To validate our intuition, we use the scenario from the previous section where the correlation for the Male group is -0.6 and for the Female group is -0.4, and then we plot the confidence scores generated by querying the same record with different sensitive attribute values and present it in Figure~\ref{fig:hexbin}.
We plot the confidence scores for the records where all predictions from various queries matched with the correct label.
In particular, we use hexagon bins to plot bivariate histograms on a 2D plane and univariate histograms on the axis margins to study the distributions of the confidence scores.
The Census19 dataset features a binary output class with labels `High Income' and `Low Income.' The sensitive attribute \texttt{MAR}, representing the marital status of the individual in the record, has two possible values: `Married' and `Single.'
In addition to `Male' and `Female,' we include the `White' group, which consists of records where the \texttt{RACE} attribute is set to `White,' with a correlation of -0.5.
High Income and Low Income records are plotted in different colors to illustrate the differences in their confidence score distributions.
The plot presents several intriguing insights.
Each distribution forms a comet-like shape with a concentrated head at the top right and a tail extending away from that point.
The direction of the tail’s trajectory differs between High Income records and Low Income records.
The trajectory features support our hypothesis; High Income records show more confidence with the sensitive attribute set to `Married' than `Single', resulting in a horizontal slant of the trajectory.
The tail trajectories of the distributions for High Income records across groups are tilted at different angles.
This occurs due to the differences in the relative frequency of sensitive attribute values, as discussed in the previous paragraph.
A similar characteristic is observed in Low Income records, except that their tail trajectories are vertically slanted.
After drawing regression lines on each set of confidence scores, the gradual shift in angle toward the center from high correlation to low correlation groups becomes even more pronounced.
Note that the regression line is slightly slanted relative to the tail’s trajectory, which occurs due to the high density of points in the top right region for High (Low) Income records.
The difference in angles between the regression lines for High Income and Low Income records also diminishes gradually from high to low correlation groups.
We term this the \emph{angular difference}, which we argue an adversary could exploit to understand and compare the correlation between groups.

The histograms clearly show that the level of correlation affects the distribution of confidence scores when records are queried with different sensitive attribute values.
An adversary could attempt to compare correlations between groups by computing the difference in confidence score distributions for records from different class labels. 
This method, however, is prone to errors because the high density shared between the distributions within the 0.95-1 confidence score range skews the calculation of differences.
We propose that computing the angular difference for each group and using it to compare correlations would be more effective, as the regression lines reflect the tail points in the distributions, which are the key differentiators.

\vspace{.2em}
\noindent\textbf{Extending to Multi-Valued/Non-Discrete Sensitive Attributes and Multi-Class Outputs}.
We draw regression lines in an $n$-dimensional space for multi-valued sensitive attributes, where $n$ denotes the number of possible sensitive attribute values.
For multi-class outputs, we generate regression lines for records from each class label and determine the average angular difference between all pairs of lines.
We propose following Mehnaz et al.~\cite{mehnaz2022your}’s approach for non-discrete sensitive attributes by binning the value ranges and using the set of bin means as a substitute for the sensitive attribute values.

\section{Attack Methodology}
\newtext{
In this section, we outline the details of the two newly proposed classes of attacks: disparity inference attack and targeted attribute inference attack. 
In the disparity inference attack, the adversary uses angular difference to rank groups based on attack vulnerability. 
In the targeted attribute inference attack, the adversary uses angular difference to strategically identify target subsets with high vulnerability and then launches attribute inference attacks on them.
}
Since both types of attacks rely on computing angular difference on groups of records, we will begin by detailing this technique.

\begin{definition}[Confidence Matrix]
\label{def:confidence_matrix}
    For a set of training data $\mathbb{D}$ containing $n$ records, a trained model $\mathcal{M}$, the confidence matrix $\mathcal{C}$ is of dimension $n \times |\mathcal{S}|$ and defined as follows -
    $$ \mathcal{C} = \left[\left[ Pr(\mathcal{M}(x')) : x' \in \mathcal{T}(x) \right]^T \quad \forall\; x \in \mathbb{D}\right]^T$$
    where $\mathcal{S}$ denotes the set of all sensitive attribute values and $\mathcal{T}(x)$ denotes the set of records generated by varying the sensitive attribute value of record $x$.
\end{definition}

\begin{definition}[Angular Difference]
\label{def:angular_difference}
    Given a training dataset $\mathbb{D}$ with $n$ records and a confidence matrix $\mathcal{C}$ generated by querying target model $\mathcal{M}$ with $\mathbb{D}$, and with $\mathbb{Y}$ representing the set of output labels, the angular difference is defined as the average difference in angle between all pairs of lines $L_{y_1}$ and $L_{y_2}$, where $y_1, y_2 \in \mathbb{Y}$. $L_y$ denotes the regression line fitted through $|\mathcal{S}|$-dimensional points from $\mathcal{C}_y$, the submatrix of $\mathcal{C}$ that contains rows corresponding to records with label $y$.
\end{definition}

\begin{algorithm}[t]
\small
    \begin{algorithmic}[1]
    \caption{Confidence Matrix Generation}
    \label{alg:confidence_matrix}
    \REQUIRE{
        $\mathcal{M}, \mathcal{N}(\mathbb{D}) \quad \triangleright$ $\mathcal{M}$ is the target model and $\mathcal{N}(\mathbb{D})$ is the non-sensitive portion of dataset $\mathbb{D}$
    }
    \ENSURE{
        $\mathcal{C}, \mathbf{t} \quad \triangleright$ $\mathcal{C}$ is the confidence matrix of the dataset $\mathbb{D}$ and $\mathbf{t}$ is a boolean vector
    }
    \FOR{each $(\mathbf{n}(x), y)$ in $\mathcal{N}(\mathbb{D})$}
        \STATE $\mathcal{X} \gets \{x' : \mathbf{n}(x) = \mathbf{n}(x') \wedge s(x') \in \mathcal{S}\}$
        \STATE $\mathcal{C}[x] \gets (Pr(\mathcal{M}(x')) : x' \in \mathcal{X})$
        \STATE $\mathbb{Y'} \gets \{M(x') : x' \in \mathcal{X}\} \quad \triangleright\;$ Set of output predicted on $\mathcal{X}$
        \IF{$\mathbb{Y'} = \{y\}$}
            \STATE $\mathbf{t}[x] \gets \texttt{true} \quad \triangleright\;$ {If all predictions were correct}
        \ELSE
            \STATE $\mathbf{t}[x] \gets \texttt{false}$
        \ENDIF
    \ENDFOR
    \RETURN $\mathcal{C}, \mathbf{t}$
    \end{algorithmic}
\end{algorithm}

\begin{algorithm}[t]
\small
    \begin{algorithmic}[1]
    \caption{Angular Difference Computation}
    \label{alg:angular_difference}
    \REQUIRE{
        $\mathcal{M}, \mathcal{N}(\mathbb{D}), \mathcal{C}, \mathbf{t} \quad \triangleright$ $\mathcal{M}$ is the target model, $\mathcal{N}(\mathbb{D})$ is the non-sensitive portion of dataset $\mathbb{D}$, $\mathcal{C}$ is the confidence matrix, and $\mathbf{t}$ is the prediction correctness vector
    }
    \ENSURE{
         $\Delta$ is the angular difference of the subset of data corresponding to group $i$
    }
    \STATE $C_y \gets \varnothing \quad \forall y \in \mathbb{Y}$
    \FOR{each $(\mathbf{n}(x), y)$ in $\mathcal{N}(\mathbb{D})$}
        \IF{$\mathbf{t}[x] = \texttt{true}$}
            \STATE $C_y \gets C_y \cup \{\mathcal{C}[x]\}$
        \ENDIF
    \ENDFOR
    \STATE $\mathcal{L} \gets \{regression\_fit(C_y) : y \in \mathbb{Y}\}$
    \STATE $\Delta \gets \underset{\underset{L_1 \neq L_2}{L_1, L_2 \in \mathcal{L}}}{\operatorname{Mean}}(angle(L_1, L_2))$
    \RETURN $\Delta$
    \end{algorithmic}
\end{algorithm}

\subsection{Computing Angular Difference}
The process of computing angular difference involves generating confidence scores from records by querying the target model while altering the sensitive attribute values. 
We call this collection of confidence scores the \emph{confidence matrix}, which is formally defined in Definition~\ref{def:confidence_matrix}.
Angular difference is formally defined in Definition~\ref{def:angular_difference}.
Algorithm~\ref{alg:confidence_matrix} outlines the steps for generating the confidence matrix and Algorithm~\ref{alg:angular_difference} outlines the steps for computing angular difference from the confidence matrix.
Initially, a set of records is generated by varying the sensitive attribute value for each original record, and the target model is queried with these sets (Lines 1-3, Algorithm~\ref{alg:confidence_matrix}). 
Subsequently, the predictions are recorded, and a boolean vector $\mathbf{t}$ tracks whether all predictions returned were correct for that record (Lines 4-8, Algorithm~\ref{alg:confidence_matrix}). 
For computing angular difference, we consider records where predictions are correct for any sensitive attribute value. 
We hypothesize that for these records, the differences in confidence scores with varying sensitive values are highly indicative of the correlation level.
We refer to $\mathbf{t}$ as the prediction correctness vector.
Afterward, for each class label, a collection of confidence scores is created from records in that class label with a prediction correctness value of \texttt{true} (Lines 1-6, Algorithm~\ref{alg:angular_difference}).
\newtext{
A regression line is then fitted to each set of confidence scores, using the dimension associated with the positive sensitive attribute value as the output. 
The angular difference is defined as the mean distance in angles between these lines (Lines 7-9, Algorithm~\ref{alg:angular_difference}).
}

To enhance readability and simplify notation, $\mathbb{D}$ will refer to the non-sensitive portion of the dataset for the remainder of this section, except when defining attack objectives, where $\mathcal{N}({\mathbb{D}})$ will be used for correctness.

\subsection{Disparity Inference Attack}
\noindent \textbf{Objective}.
Let $\mathcal{M}$ be a target model trained on dataset $\mathcal{N}(\mathbb{D})$, which can be divided into $k$ non-overlapping subsets $\mathcal{N}(\mathbb{D}_1), \mathcal{N}(\mathbb{D}_2), \dots,$ $ \mathcal{N}(\mathbb{D}_k)$. 
The success rate of an attack $\mathcal{A}$ on target model $\mathcal{M}$ using dataset $\mathbb{D}$ is denoted by $ASR(\mathcal{M},\;\mathcal{N}(\mathbb{D}),\;\mathcal{A})$, with $\mathcal{N}(\mathbb{D})$ indicating the non-sensitive part of the data accessible to the adversary.
The attack vulnerability ranking $\mathcal{R} = (r_1, r_2, \dots, r_k)$ of the groups corresponding to subsets $\mathbb{D}_1$, $\mathbb{D}_2$, \dots, $\mathbb{D}_k$ is defined such that:
\begin{align*}
    \{r_1, r_2, \dots, r_k\} &= [1, k] \\
    ASR(\mathcal{M},\;\mathcal{N}(\mathbb{D}_{r_i}),\;\mathcal{A}) &\geq ASR(\mathcal{M},\;\mathcal{N}(\mathbb{D}_{r_j}), \mathcal{A}) \\
    &\forall\; 1 \leq i < j \leq k
\end{align*}
The attacker’s goal is to find $\mathcal{R}$ or a ranking very close to $\mathcal{R}$.
It is important to consider that the attacker does not know the true sensitive attribute values and therefore cannot determine the exact $ASR(\mathcal{M}, \mathcal{N}(\mathbb{D}_i), \mathcal{A})$ for all $i \in [1, k]$. 
Otherwise, the attacker’s goal would be trivial.

\textbf{Attack Steps}.
Query $\mathcal{M}$ using Algorithm~\ref{alg:confidence_matrix} to generate the confidence matrix $\mathcal{C}$ for $\mathbb{D}$. 
Next, use Algorithm~\ref{alg:angular_difference} to calculate the angular difference $\Delta_i$ for each subset $\mathbb{D}_i$. 
Rank the indices $1, 2, \dots, k$ by decreasing $\Delta_i$ values and output this as the desired ranking.

\subsection{Targeted Attribute Inference Attack}
\textbf{Objective}.
Let $\mathcal{M}$ be a target model trained on dataset $\mathbb{D}$.
The attack success rate of an attack $\mathcal{A}$ on target model $\mathcal{M}$ using dataset $\mathbb{D}$ is denoted by $ASR(\mathcal{M},\;\mathcal{N}(\mathbb{D}),\;\mathcal{A})$ where $\mathcal{N}(\mathbb{D})$ denotes the non-sensitive portion of the data available to the adversary.
The objective of the adversary is to find a dataset $\mathbb{D}_{target} \subset \mathbb{D}$ satisfying the following conditions:
\begin{equation}
\begin{aligned}
\left| \frac{|\mathbb{D}_{target}|}{|\mathbb{D}|} - \kappa \right| < \epsilon
\end{aligned}
\label{eq:targeted_attack_size_condition}
\end{equation}
\begin{equation}
\begin{aligned}
ASR(\mathcal{M},\;\mathcal{N}(\mathbb{D}_{\text{target}}),\;\mathcal{A}) &\geq ASR(\mathcal{M},\;\mathcal{N}(\mathbb{D}'),\;\mathcal{A}) \\
\forall\;\mathbb{D}' \in \{\mathbb{D}' \subset \mathbb{D} &\mid |\mathbb{D}'| > |\mathbb{D}_{\text{target}}|\}
\end{aligned}
\label{eq:targeted_attack_condition}
\end{equation}
where $\kappa$ is the attack budget controlling the target dataset size within $0$ to $0.5$.
The value of $\epsilon$ is set to a very small amount to ensure that the size of the target dataset meets the attack budget.
Condition~\ref{eq:targeted_attack_condition} ensures that the adversary performs better in the target subset than in any subset of equal or greater size.
The degree of disparate vulnerability determines how much performance improvement the adversary can obtain through a targeted attack versus an untargeted attack.

\textbf{Key Insights}.
It is computationally intractable to evaluate attack success rate on all possible $\mathbb{D}'$ to find a $\mathbb{D}_{target}$ that satisfies Condition~\ref{eq:targeted_attack_condition}.
However, we can constrain our exploration to the subsets of $\mathbb{D}'$ defined by restricting one or more of non-sensitive attributes of the records to a subset of their respective possible values.
Our next challenge is finding a method to compare attack success rates between two subsets without using an auxiliary dataset, since our threat model assumes the adversary lacks access to any additional data.
Similar to our previous attacks, we propose utilizing the strong connection between angular difference and attack success rate on a subset.
Hence, the adversary's goal shifts to finding the subset that has the greatest angular difference.
Although constraining the number of subsets to explore makes the adversary's objective feasible, it is still exponentially costly to naively compute angular difference on the constrained set of subsets.
Therefore, we propose two innovative strategies to optimally explore the subset space which are outlined in detail in the sections that follow.

\subsubsection*{Single Attribute-based Targeted Attack}
\label{sec:single_attributed_based_targeted_attack}
\begin{enumerate}[leftmargin=0pt, itemindent=15pt]
    \item Randomly sample ${\mathbb{D}^q}$ from $\mathbb{D}$ with the condition: $|{\mathbb{D}^q}| = q \times |\mathbb{D}|$ where $q$ is the query budget to limit the number of queries made to the model.
    \item Use Algorithm \ref{alg:confidence_matrix} to query $\mathcal{M}$ and generate confidence matrix $\mathcal{C}$ on $\mathbb{D}^q$.
    \item For each non-sensitive attribute $a$, split ${\mathbb{D}^q}$ into subsets $\mathbb{D}^q_1, \mathbb{D}^q_2, \dots, \mathbb{D}^q_k$ such that $\mathbb{D}^q_i = \{ x \in \mathbb{D}^q\ | a(x) = \mathcal{A}_i \}$ for all $i \in [1, k]$ with $\mathcal{A}$ representing a set of size $k$ that contains possible values of $a$ and $a(x)$ denoting the value assigned to attribute $a$ in record $x$. Compute the angular difference on each subset $\mathbb{D}^q_i$ and put them in the vector ${\Delta}_a^q$. Define, $\mathrm{range({\Delta}_a^q)} = \max{({\Delta}_a^q)} - \min{({\Delta}_a^q)}$.
    \item Find the attribute $a$ with the highest $\mathrm{range({\Delta}_a^q)}$. Let, $\mathbb{D}_i = \{ x \in \mathbb{D}\ | a(x) = \mathcal{A}_i \}$ and $\Delta_i$ be the angular difference on subset $\mathbb{D}^q_i$ that we computed on the previous step for all $i \in [1, k]$.
    \item Rank indices $1, 2, \dots, k$ based on the value of $\Delta_i$ in increasing order. Let, $\mathbb{I}$ denote this ordered set of indices and $\mathbb{I}_{[m, n]}$ denote the ordered subset of $\mathbb{I}$ starting at index $m$ and ending at index $n$.
    \item Let, $\mathbb{D}_{[1, m]} = \bigcup_{i \in \mathbb{I}_{[m, n]}}{\mathbb{D}_i}$ for all $m \in [1, k]$. Output $\mathbb{D}_{[1, m]}$ that satisfies Condition~\ref{eq:targeted_attack_size_condition}.
\end{enumerate}
The initial four steps identify the most suitable single attribute for exploring the subset space. 
Intuitively, the attribute with the widest range of angular differences in its subsets is likely to contain the most vulnerable subsets among those defined by a single attribute.
Then the attacker can rank the vulnerability of subsets defined by the selected attribute by ranking corresponding angular difference values (step 5) and aggregating the most vulnerable subsets in decreasing order of angular difference until the attack budget is reached (step 6).

\subsubsection*{Nested Attribute-based Targeted Attack}
\label{sec:nested_attribute_based_targeted_attack}
\begin{enumerate}[leftmargin=0pt, itemindent=15pt]
    \item Set the depth or the number of nested attributes $d$ to $\lceil \log_2({\kappa}) \rceil$.
    \item Follow steps 1 to 3 exactly as specified in the single attribute-based targeted attack (section \ref{sec:single_attributed_based_targeted_attack}) to find the range of angular differences for each attribute.
    \item Select the attributes $a_1, a_2, .., a_d$ that are among the top-$d$ in terms of the largest ranges of angular differences. Let $\mathbb{D}_j^i$ denote the set of records with the attribute $a_i$ set to the $j$-th value from $\mathcal{A}_i$.
    \item For each attribute $a_i \in \{a_1, a_2, .., a_d\}$, define $\mathbb{I}^i$ as the ordered set of indices ranked on angular differences of groups defined by attribute $a_i$. Let $\mathbb{I}^i_m$ denote the top-$m$ groups defined by attribute $a_i$ in terms of their angular difference. 
    \item Let the \emph{above-average-risk segment} denote the set of the most vulnerable groups based on their angular differences, ensuring they comprise close to half of the total records in all groups defined by the attribute $a_i$. For each attribute $a_i \in \{a_1, a_2, .., a_d\}$, define $\mathbb{I}^i_{1/2}$ as the indices within $\mathbb{I}^I$ that identify the groups constituting the above-average-risk segment. Formally, $\mathbb{I}^i_{1/2} = \underset{m \in [1, |\mathcal{A}_i|]}{argmin} \left| \left| \cup_{j \in \mathbb{I}^i_m} \mathbb{D}_j^i \right| / |\mathbb{D}| - 0.5 \right|$. Find above-average-risk segment, $\mathbb{D}^i_{1/2} = \cup_{j \in \mathbb{I}^i_{1/2}} \mathbb{D}_j^i$, for each attribute $a_i \in \{a_1, a_2, .., a_{d-1}\}$. 
    \item Let, $\mathbb{D}_{m} = \mathbb{D}^1_{1/2} \bigcap \dots \bigcap \mathbb{D}^{d-1}_{1/2} \bigcap \left( \bigcup\limits_{i \in \mathbb{I}^d_m} \mathbb{D}_i^d \right)$ for $m$ in $[1, |\mathcal{A}_d|]$. Output $\mathbb{D}_{m}$ with the lowest value of $m$ that satisfies Condition~\ref{eq:targeted_attack_size_condition}.
\end{enumerate}

In this strategy, we examine the subset space made up of nested groups, which are intersections of groups defined by different attributes.
We achieve this by combining the above-average-risk segment from each attribute and forming nested groups through their intersections.
To comply with the attack budget, we cap the number of attributes considered at $d$.
This greedy approach is adopted due to the exponential computational complexity of considering every combination of nested groups.  
Steps 1 to 3 involve identifying the best $d$ attributes in terms of the range of angular differences. 
The attack budget may not permit selecting the full above-average-risk segment from the last chosen attribute; therefore, for that attribute, we select as many groups as possible while satisfying Condition~\ref{eq:targeted_attack_size_condition} as shown in step 6.

\section{Experiments}
In this section, we explain our experimental arrangement, datasets, machine learning models, and performance metrics. 
We then examine the performance of our proposed attacks.

\subsection{Experimental Setup}
\label{sec:experimental_setup}
\textbf{Datasets.}
We use the following three datasets in our experiments:
Census19~\cite{censusPUMSData}, Texas-100X~\cite{Texas}, and Adult~\cite{misc_adult_2}.
More details of the datasets are presented in Appendix~\ref{app:datasets}.

\noindent\textbf{Sampling Technique.}
Both Census19 and Texas-100X datasets contain around a million records each, making them ideal candidates for selective sampling to manage dataset-specific variables, including the relationship between sensitive and output attributes.
In contrast, earlier studies~\cite{fredrikson2015model, mehnaz2022your} utilizing datasets like Adult~\cite{misc_adult_2} and GSS~\cite{norcGSSGeneral} struggled with limited data sizes, barely sufficient for training a model that could achieve reliable accuracy on a test set and could not conduct detailed sampling. 
In our experimental evaluation, we incorporate a sampling technique that allows us to set a predefined correlation between the sensitive and output attributes not only for the full training data but also for specific groups.
The specifics of this sampling technique are detailed step-by-step below. 
\begin{enumerate}
    \item Let $n$ denote the number of desired samples for the training data or a specific group. Additionally, let $m$ denote the ratio between the number of samples with negative sensitive values and the number of samples with positive sensitive values, and let $c$ denote the desired correlation between the sensitive attribute and the output. 
    \item Let $n_+^+$ and $n_+^-$ denote the number of samples to be picked that have positive sensitive values with positive and negative output values, respectively. Similarly let, $n_-^+$ and $n_-^-$ denote the number of samples to be picked that have negative sensitive values with positive and negative output values, respectively. Simply put, the subscript denotes the sensitive value and the superscript denotes the output value.
    $n_-^+$ = $\lfloor \frac{\sqrt{m} \times (\sqrt{m} - c) \times n}{2 \times (m+1)} \rfloor$,
    $n_-^-$ = $\lfloor \frac{\sqrt{m} \times (\sqrt{m} + c) \times n}{2 \times (m+1)} \rfloor$\\
    $n_+^-$ = $\lceil \frac{n}{2} - n_-^- \rceil$,  $n_+^+$ = $\lceil \frac{n}{2} - n_-^+ \rceil$
\end{enumerate}
\newtext{
The sampling method used above ensures that the desired correlation is achieved while maintaining a balanced number of positive and negative output samples.
The correctness proof of our claim is provided in Appendix~\ref{sec:correctness_proof}.
For all experiments on Census19 and Texas-100X, the value of $m$ is set to 1 to align with the sensitive attribute distribution of the original dataset.
This sampling method enables precise control over attribute correlations, which enhances our capacity to identify significant trends or effects on attack efficacy, while still reflecting real-life data—something that prior research methodologies have not been able to achieve.
In addition, we experiment with the Adult dataset without using controlled sampling to create an even more realistic setting.
}

\noindent\textbf{Model Training.}
For our experiments with both Texas-100X and Census19 datasets, we select 50,000 records to create the training set for training the neural network model. 
Additionally, we randomly sample another 50,000 records from the remaining data to form the test set, ensuring that the training and test sets are mutually exclusive.
\newtext{
For the Adult dataset, we utilize the full dataset, splitting it into training and test data as done in~\cite{mehnaz2022your}.
}
We use Scikit-learn's~\cite{scikit-learn} implementation of Multi-layer Perceptron (MLP) as our default class of machine learning model.
The architecture and training hyperparameter details are put into Appendix~\ref{appendix:model_training}.

\noindent\textbf{Code}.
We have released the full codebase, including all scripts and model checkpoints necessary for replicating our experiments, via a public repository: \url{https://zenodo.org/records/14732956}. For the most accurate and current implementation, please refer to the latest version.

\subsection{Ideal vs. Practical Imputation Attacks}
\label{sec:ideal_vs_practical_imputation}
\newtext{
An ideal imputation attack is characterized by the adversary having an auxiliary dataset that precisely matches the distribution of the target data. 
This includes having similar distribution properties, such as correlation and marginal priors, for any group of records.
However, it is impractical to assume that an adversary, who is typically external to the organization owning the private training data, would be able to acquire an auxiliary dataset that accurately mirrors the distribution at a granular level.
Thus, any dataset obtained by a realistic adversary would differ in distributional properties, either at a macro-level (calculated across the whole dataset) or at a micro-level (calculated for specific subsets within the dataset). 
An imputation attack conducted with such auxiliary data is termed a practical imputation attack.
In this section, we perform two experiments on Adult dataset to evaluate the performance of practical imputation attacks against ideal imputation attacks, highlighting the substantial gap between them.
The first experiment explores dataset-level distributional drift in the auxiliary dataset by altering the marginal prior relative to the training dataset. 
The second experiment investigates group-level distributional drift in the auxiliary dataset, maintaining the same dataset-level correlation as the training data but with different group-level correlations.
}

\setlength{\tabcolsep}{3pt}

\begin{figure}[t]
    \centering
    \begin{minipage}{0.37\columnwidth}  
        \centering
        \vspace{0.3cm}
        \resizebox{\linewidth}{!}{%
            \begin{tabular}{l|rrrr}
            \multirow{2}{*}{$\eta$} & \multicolumn{4}{c}{$|D_{aux}|$} \\
             & 100 & 500 & 1000 & 5000 \\
            \hline
            0.1 & 47.96 & 54.41 & 60.15 & 62.54 \\
            0.2 & 57.79 & 68.95 & 66.93 & 67.68 \\
            0.3 & 64.16 & 71.50 & 70.58 & 71.57 \\
            0.4 & 71.47 & 71.98 & 72.89 & 73.31 \\
            0.5 & 68.84 & 73.43 & 71.82 & 73.46 \\
            \bottomrule
            \end{tabular}
        }
        \caption*{(a)}
        \label{fig:dataset_level_distrib_drift}
    \end{minipage}
    \hfill
    \begin{minipage}{0.6\columnwidth}  
        \centering
        \includegraphics[width=\linewidth]{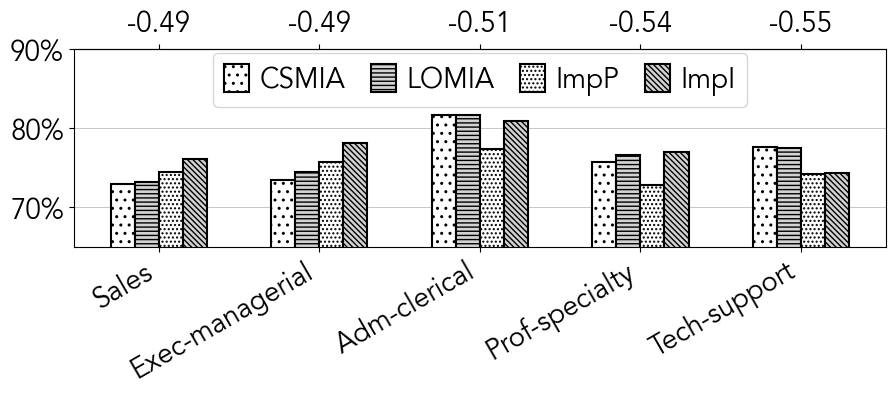}
        \caption*{(b)}
    \end{minipage}
    \caption{(a) Imputation attack performance across 2 dimensions: Auxiliary Dataset Size (row) and Marginal Prior, $\eta$ (column). (b) Imputation Attacks (Practical -- ImpP, Ideal -- ImpI) versus AI attacks (CSMIA, LOMIA) in Occupation Groups with high correlation. Bottom x-axis labels indicate group Names; Top x-axis labels indicate Group Correlation Values; Y-axis denotes Attack Accuracy.}
    \label{fig:distrib_drift}
\end{figure}

\textbf{Dataset-level Distributional Drift}.
In this experiment, we vary the marginal prior $\eta$ of the auxiliary dataset, which represents the fraction of positive samples, from 0.5 to 0.1, thus deviating it further from the training data’s marginal prior of 0.52.
We also consider auxiliary datasets of various sizes, ranging from 5000 to 100.
Figure~\ref{fig:distrib_drift}(a) presents the performance of imputation attacks for every combination of auxiliary datasets across the two analyzed dimensions.
The results demonstrate a significant performance decline with increased deviation in $\eta$. 
Notably, for $\eta$ values of 0.1 and 0.2, the imputation attack’s performance falls below that of CSMIA (69.97) and LOMIA (70.61) regardless of auxiliary dataset size.
Imputation attacks outperform CSMIA and LOMIA only when the auxiliary dataset’s marginal prior is close to the original. 
This implies that \emph{practical imputation attacks may only achieve high performance when conditions are nearly ideal. Otherwise, they are likely to perform worse than existing AI attacks.}

\newtext{
\textbf{Group-level Distributional Drift}.
In this experiment, we use an auxiliary dataset where the correlation for each occupation group is -0.44, which matches the overall correlation of -0.4412 in the training data. 
However, correlations within the groups of the original training data vary significantly, ranging from -0.17 to -0.55, suggesting a group-level distributional drift in the auxiliary data.
We perform an imputation attack with this auxiliary dataset, an ideal imputation attack, LOMIA, and CSMIA, and display the attack success rate of all attacks for groups with higher-than-overall correlation in Figure~\ref{fig:distrib_drift}(b).
According to the results, CSMIA and LOMIA outperform the practical imputation attack in the top 3 out of 5 vulnerable groups. 
By contrast, ideal imputation attack performance is very similar to CSMIA and LOMIA in 2 out of these 3 groups.
Nevertheless, the results indicate that \emph{a practical imputation attack with group-level distributional drift is likely to perform poorly in highly vulnerable groups compared to attribute inference (AI) attacks.}
}

\newtext{
The results of the experiments above suggest that practical imputation attacks, the only type that a realistic adversary can execute, are likely to underperform relative to AI attacks despite having access to an auxiliary dataset that the AI attacks do not assume.
This implies that AI attacks on ML models enable adversaries to more accurately predict sensitive attributes compared to practical imputation attacks, signaling privacy leakage from these models. 
In short, even if AI attacks do not surpass ideal imputation attacks, their superiority over practical imputation attacks demonstrates privacy leakage from ML models.
Thus, practical imputation attacks should be viewed as a baseline for AI attack evaluation, whereas ideal imputation attacks can provide a useful benchmark for assessing privacy leakage.
In the sections that follow, we first evaluate the effectiveness of our Disparity Inference Attack and then we investigate the extent of privacy leakage resulting from the two targeted AI attack types.
}

\subsection{Disparity Inference Attack Performance}
\label{sec:exploiting_disparate_vulnerability}
\textbf{Sensitive Attribute and Group Attribute Selection}.
We select the \texttt{MAR} column, denoting marital status, as the sensitive attribute for the Census-19 dataset, and for the Texas-100X dataset, we pick \texttt{SEX\_CODE} as the sensitive attribute to be inferred.
For the Census-19 dataset, we employ the \texttt{State} attribute to divide the training data into 51 groups and for the Texas-100X dataset, we use the \texttt{PAT\_STATUS} attribute to divide the training data into 10 groups.
Since attribute inference attack performance is significantly influenced by the correlation between the sensitive attribute and output, we assign different correlation values to each of the 51 groups to achieve varying levels of vulnerability. 
Each state is indexed from 0 to 50, and the desired correlation between the sensitive and output attributes is set to $-0.01 \times i$ for records from the state with index $i$. 
Whenever possible, 1,000 records are sampled per group, or fewer if fewer are available, while preserving the specified correlation.
Similarly for Texas-100X, each group, defined by a specific value of \texttt{PAT\_STATUS}, is assigned a desired correlation value from 0, 0.05, 0.10, \dots, 0.45.
Afterward, we train target models on each subset of records.

\begin{figure}
    \centering
    \begin{minipage}{0.23\textwidth}
         \includegraphics[width=\linewidth]{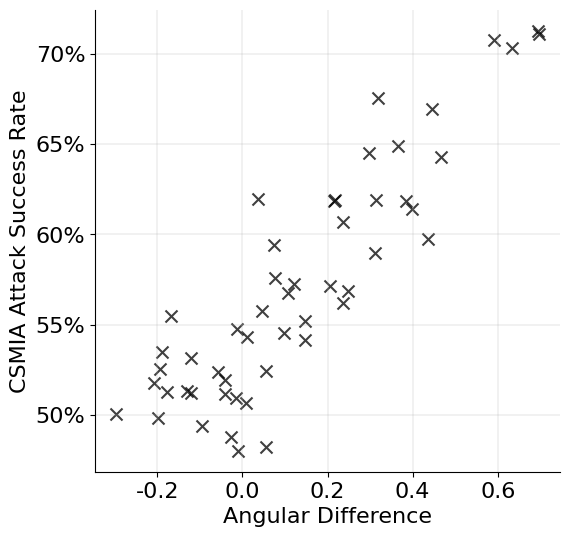}
          \label{fig:disparity_inference_ang_diff_vs_attack_perf_CSMIA}
    \end{minipage}
    \begin{minipage}{0.23\textwidth}
        \includegraphics[width=\linewidth]{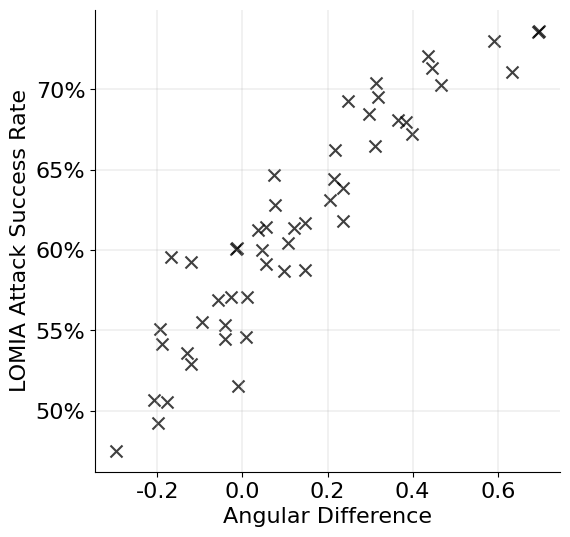}
        \label{fig:disparity_inference_ang_diff_vs_attack_perf_LOMIA}
        \end{minipage} 
\caption{Angular difference vs. attack performance (CSMIA - left, LOMIA - right) of 51 states from Census19 dataset. Accuracy is used as attack performance metric.}
\label{fig:disparity_inference_ang_diff_vs_attack_perf}
\end{figure}

\textbf{Evaluation}.
We launch our proposed disparity inference attack, which computes angular differences for each group and ranks them based on that.
Figure~\ref{fig:disparity_inference_ang_diff_vs_attack_perf} plots the angular difference in X-axis and the actual attack performance in Y-axis for both CSMIA (left plot) and LOMIA (right plot).
\newtext{
The results clearly demonstrate a strong association between the angular difference and the success rate of the attack. 
Specifically, a higher angular difference for a group indicates that both attacks will likely perform well, whereas a lower angular difference suggests that the performance will be poor.
}
\emph{Therefore, an adversary can compute angular differences on multiple groups and use that as an indicator of the quality of attack predictions made on the records of the groups}.

\setlength{\tabcolsep}{4.5pt}
\begin{table*}[t]
    \centering
    \tiny
    \begin{tabular}{ccccc}
        & \multicolumn{4}{c}{Census19} \\
        \hline
        & \multicolumn{2}{c}{Kendall Tau} & \multicolumn{2}{c}{Spearman R} \\
        \cline{2-5}
        & CSMIA & LOMIA & CSMIA & LOMIA \\
        \hline
        Disparity Inference Attack & 0.6914 (1.39e-12) & 0.7579 (8.30e-15) & 0.8767 (7.23e-17) & 0.9104 (5.06e-20) \\
        Baseline & -0.0759 (4.37e-01) & -0.0931 (3.40e-01) & -0.1225 (3.97e-01) & -0.1275 (3.77e-01) \\
        \hline
    \end{tabular}
    \hfill
    \begin{tabular}{ccccc}
    & \multicolumn{4}{c}{Texas-100X} \\
    \hline
    & \multicolumn{2}{c}{Kendall Tau} & \multicolumn{2}{c}{Spearman R} \\
    \cline{2-5}
    & CSMIA & LOMIA & CSMIA & LOMIA \\
    \hline
    Disparity Inference Attack & -0.7778 (9.46e-04) & -0.7778 (9.46e-04) & -0.9273 (1.12e-04) & -0.9152 (2.04e-04) \\
    Baseline & -0.2444 (3.81e-01) & -0.2444 (3.81e-01) & -0.3576 (3.10e-01) & -0.3939 (2.60e-01) \\
    \hline
\end{tabular}
    \caption{Comparative evaluation of the ranking quality of disparity inference attacks versus baseline attacks employing an auxiliary dataset. The values inside the parentheses are p-values corresponding to the null hypothesis.}
    \label{tab:disparity_inference_vs_baseline}
\end{table*}

\newtext{
To evaluate the ranking quality of our disparity inference attack, we use two statistical metrics: Kendall’s Tau~\cite{kendall1938new} and Spearman’s Rank Correlation~\cite{spearman1904proof}.
}
These two metrics, known for their robustness against outliers and ability to minimize the impact of extreme values, have been extensively applied across numerous scientific domains.
The metrics range from -1 to 1, however, ranking performing close to 0 is considered poor while ranking performance further from 0 (close to either -1 or 1) is considered good.
That is because a ranking performing negatively implies the order is in reverse and may not necessarily be lacking in utility.
Furthermore, we consider the following baseline vulnerability ranking attack for comparison -- we assume that the attacker possesses an auxiliary dataset that is the same size as the training dataset in which the sensitive attribute values of all records are known.
\newtext{
To ensure that the auxiliary dataset and the original dataset have different distributions, we randomly sample from the full versions of Census19 and Texas-100X to create the auxiliary dataset.
}
The attacker launches CSMIA on the auxiliary dataset and evaluates attack performance on all the groups and ranks them based on the attack performance on the auxiliary data.
Table~\ref{tab:disparity_inference_vs_baseline} presents the comparative evaluation of ranking quality between our proposed disparity inference attack and the baseline attack we considered.
The results show that the vulnerability ranking by our disparity inference attack is far superior to that of the baseline attack.
\emph{This not only establishes that the task of disparity inference is not trivial but also our proposed approach is very effective in ranking the groups in terms of their vulnerability.}
The extremely low p-values corresponding to the null hypothesis indicate that the closeness between the rankings is not due to chance and the null hypothesis can be rejected.

\subsection{Targeted Attribute Inference Attack}

\setlength{\tabcolsep}{5.5pt}
\begin{table*}[t]
    \centering
    \tiny
    \hfill
    \begin{tabular}{crrrrrrr}
        \multicolumn{8}{c}{Census19} \\
        \hline
       $\kappa$  & 1 & 0.75 & 0.5 & 0.375 & 0.25 & 0.1 & 0.05  \\
        \hline
        {CSMIA} & 62.56 & 64.73 & 67.43 & 69.02 & 70.42 & 72.54 & 73.27 \\
        {LOMIA} & 61.24 & 63.71 & 67.56 & 69.45 & 70.82 & 72.92 & 73.78 \\
        {ImpI} & 65.38 & 65.65 & 66.10 & 65.91 & 65.71 & 67.30 & 66.85 \\
        {ImpP} & 62.99 & 63.30 & 63.57 & 62.96 & 64.12 & 64.17 & 64.25 \\
        \hline
    \end{tabular}
    \hfill
    \begin{tabular}{crrrrr}
        \multicolumn{6}{c}{Texas-100X} \\
        \hline
       $\kappa$  & 1 & 0.75 & 0.5 & 0.25 & 0.1  \\
        \hline
        {CSMIA} & 60.95 & 62.82 & 64.39 & 61.00 & 62.82 \\
        {LOMIA} & 61.50 & 63.90 & 66.72 & 64.68 & 69.68 \\
        {ImpI} & 59.33 & 63.95 & 65.01 & 65.57 & 66.75 \\
        {ImpP} & 51.64 & 47.55 & 46.73 & 46.26 & 48.97 \\
        \hline
    \end{tabular}
    \hfill
    \begin{tabular}{crrrrr}
        \multicolumn{6}{c}{Adult} \\
        \hline
       $\kappa$  & 1 & 0.75 & 0.5 & 0.25 & 0.1  \\
        \hline
        {CSMIA} & 69.96 & 69.14 & 72.37 & 74.83 & 81.61 \\
        {LOMIA} & 70.61 & 69.79 & 73.36 & 74.86 & 81.68 \\
        {ImpI} & 74.46 & 76.96 & 77.72 & 79.82 & 81.52 \\
        {ImpP} & 65.05 & 68.31 & 69.70 & 63.94 & 65.21 \\
        \hline
    \end{tabular}
    \hspace{15pt}
    \caption{Attack success rate of single attribute-based targeted inference attack compared with targeted imputation baselines. Accuracy is used as the metric for attack success rate.}
    \label{tab:targeted_inference_attack_by_state}
\end{table*}

\setlength{\tabcolsep}{5.5pt}
\begin{table*}[t]
    \centering
    \tiny
    \begin{tabular}{crrrrr}
        \multicolumn{6}{c}{Census19} \\
        \hline
       $\kappa$  & 1 (0) & 0.5 (1) & 0.375 (2) & 0.25 (3) & 0.1 (4)  \\
        \hline
        {CSMIA} & 62.56 & 67.43 & 67.43 & 69.70 & 69.36 \\
        {LOMIA} & 61.24 & 67.56 & 66.96 & 68.20 & 70.26 \\
        {ImpI} & 65.38 & 66.06 & 65.98 & 66.17 & 66.85 \\
        {ImpP} & 62.99 & 63.57 & 63.42 & 63.60 & 60.86 \\
        \hline
    \end{tabular}
    \hfill
    \begin{tabular}{crrrrrr}
        \multicolumn{7}{c}{Texas-100X} \\
        \hline
       $\kappa$  & 1 (0) & 0.5 (1) & 0.25 (2) & 0.1 (3) & 0.05 (4) & 0.01 (5) \\
        \hline
        {CSMIA} & 60.95 & 64.39 & 61.77 & 66.12 & 67.77 & 100.00 \\
        {LOMIA} & 61.50 & 66.72 & 63.64 & 64.85 & 66.19 & 100.00 \\
        {ImpI} & 59.33 & 63.97 & 63.78 & 72.02 & 72.67 & 78.42 \\
        {ImpP} & 51.64 & 46.73 & 48.11 & 45.38 & 48.40 & 43.15 \\
        \hline
    \end{tabular}
    \hfill
    \begin{tabular}{crrrrr}
        \multicolumn{6}{c}{Adult} \\
        \hline
       $\kappa$  & 1 (0) & 0.5 (1) & 0.375 (2) & 0.25 (3) & 0.1 (4)  \\
        \hline
        {CSMIA} & 69.97 & 71.18 & 73.72 & 77.57 & 86.74 \\
        {LOMIA} & 70.61 & 72.19 & 73.90 & 73.90 & 86.77 \\
        {ImpI} & 74.46 & 77.72 & 76.47 & 78.16 & 77.86 \\
        {ImpP} & 65.05 & 69.70 & 73.59 & 58.89 & 68.65 \\
        \hline
    \end{tabular}
    \caption{Attack performance of nested attribute-based targeted inference attacks compared with targeted imputation baselines. The values inside parentheses denote the number of nested attributes considered. Accuracy is used as the attack performance metric.}
    \label{tab:nested_attribute_based_targeted_ai_results}
\end{table*}

We adopt the same configuration as the disparity inference attack for our evaluation.
\newtext{
Alongside Census19 and Texas-100X, we also perform targeted attribute inference (AI) experiments on the full Adult dataset to illustrate the real-world effects of the attacks.
The adversary follows the steps outlined in Section~\ref{sec:single_attributed_based_targeted_attack}\reviseddeletetwo{ and Section~\ref{sec:nested_attribute_based_targeted_attack}}, using either CSMIA or LOMIA as the underlying attribute inference algorithm.
As baselines, we consider two variations of the imputation attack: ImpI and ImpP. 
ImpI uses an auxiliary set that matches the distribution of the original data, while ImpP uses an auxiliary set with a different distribution.
For both methods, the adversary applies an imputation attack on the auxiliary data to determine the attack success rate across various groups. 
Using the success rate as a substitute for angular difference, the adversary then mimics the steps of our proposed targeted attacks to initiate a targeted imputation attack.
Once the target subset is chosen, the adversary carries out an imputation attack on it.
}

\textbf{Single Attribute-based Targeted AI}.
Table~\ref{tab:targeted_inference_attack_by_state} presents the evaluation results of the single attribute-based targeted inference attack.
\newtext{
To evaluate the attack, we use various $\kappa$ values from 1 to 0.1. 
For Census19 specifically, we set $\kappa$ as low as 0.05, which is feasible because the grouping attribute has 51 unique values.
$\kappa = 1$ denotes the untargeted version.
For Census19, Texas-100X, and Adult, we choose \texttt{STATE}, \texttt{PAT\_STATUS}, and \texttt{Occupation} respectively as the grouping attributes, since these exhibit the greatest range in angular difference.
The results show that targeted attacks using both CSMIA and LOMIA consistently improve in accuracy as $\kappa$ is reduced, demonstrating a clear trend across nearly all cases.
The CSMIA variant targeted attack results in performance increases of 17.12\% for Census19, 5.65\% for Texas-100X, and 16.66\% for Adult, compared to the untargeted counterpart. For the LOMIA variant, the performance gains are 20.48\%, 13.31\%, and 15.68\%, respectively.
}

\newtext{
\textbf{Nested Attribute-based Targeted AI}.
Table~\ref{tab:nested_attribute_based_targeted_ai_results} presents the evaluation results of the nested attribute-based targeted inference attack.
The top-d attribute ordered set selected for Census19, Texas-100X, and Adult are \{\texttt{STATE}, \texttt{SCHOOL}, \texttt{RACE}, \texttt{SEX}\}, \{\texttt{PAT\_STATUS}, \texttt{RACE}, \texttt{ADMITTING\_DIAGNOSIS}, \texttt{TYPE\_OF\_ADMISSION}, \texttt{SOURCE\_OF\_ADMISSION}\}, \{\texttt{Occupation}, \texttt{Work}, \texttt{Race}, \texttt{Sex}\} respectively.
Similar to single attribute-based attacks, nested attribute-based targeted AI shows increased attack performance with a smaller attack budget, i.e., increased depth of attributes.
For the Adult dataset, the attack success rate increases up to $86.77\%$, and for the Texas-100X dataset, it reaches $100\%$ at a depth of 5 for both CSMIA and LOMIA variants.
}

\newtext{
\textbf{Comparison with Baselines}.
The pattern of improved performance as $\kappa$ decreases is not observed for ImpP, implying that when an adversary’s auxiliary dataset differs in distribution, a targeted imputation attack will not outperform an untargeted attack.
ImpI demonstrates improved performance with decreasing $\kappa$ values, though at a slower rate than our proposed attack, suggesting a correlation between group-level attack success rates on ImpI’s auxiliary dataset and the original data.
Given that obtaining auxiliary dataset with the same distribution as the private training data is impractical, our proposed targeted AI attacks remain the most feasible option for adversary with minimal knowledge and capabilities.
}

\setlength{\tabcolsep}{5.5pt}
\begin{table}[t]
    \centering
    \tiny
    \begin{tabular}{ccrrrrrr}
        \hline
        \multicolumn{2}{r}{Number of MLP Layers $\rightarrow $} & \multicolumn{2}{c}{4} & \multicolumn{2}{c}{3} & \multicolumn{2}{c}{2} \\
        \hline
            & \multicolumn{1}{r}{$\kappa \rightarrow$}  & 1 & 0.05 & 1 & 0.05 & 1 & 0.05  \\
        \hline
        \multirow{2}{*}{Single Attribute-based} &  {CSMIA} & 63.82 & 72.42 & 62.56 & 73.27 & 60.73 & 71.85  \\
         & {LOMIA} & 61.63 & 73.02 & 61.24 & 73.78 & 60.79 & 73.66 \\
        \hline
            & \multicolumn{1}{r}{$\kappa \rightarrow$}  & 1 & 0.1  & 1 & 0.1 & 1 & 0.1   \\
        \hline
        \multirow{2}{*}{Nested Attribute-based} &  {CSMIA} & 63.82 & 68.66 & 62.56 & 69.36 & 60.73 & 68.96  \\
         & {LOMIA} & 61.63 & 70.75 & 61.24 & 70.26 & 60.79 & 70.56 \\
        \hline
    \end{tabular}
    \caption{Attack success rate of targeted inference attacks across MLP models of varying depths.}
    \label{tab:targeted_ai_performance_across_mlp_layers}
\end{table}

\textbf{Effect of MLP Architecture}.
Targeted AI attacks are launched on MLPs with varying depths (with hidden layers ranging from 2 to 4) trained on Census19 to examine whether the complexity of ML models influences their susceptibility to AI attacks. 
Table~\ref{tab:targeted_ai_performance_across_mlp_layers} displays the experimental results. 
The findings indicate that both single and nested variations show similar performance across MLPs with different numbers of layers, suggesting that the architecture of MLPs does not influence vulnerability to AI attacks.

\section{Potential Mitigation Strategies}
In this section, we examine potential strategies to mitigate disparate vulnerabilities that an adversary could exploit to launch various types of attacks, as demonstrated in the previous section.
Existing literature~\cite{wang2021improving, jayaraman2022attribute} provides techniques to defend against attribute inference attacks, but none propose methods to mitigate disparity between groups.
Dibbo et al.~\cite{dibbo2023model} apply defense techniques tailored to mitigate disparity in other domains~\cite{kulynych2022disparate} or to defend against attribute inference attacks without attempting to address disparity~\cite{wang2021improving}, finding limited success in mitigating disparity against attribute inference attacks.
We investigate two defensive strategies: the first adapts the mutual information regularization (MIR) technique~\cite{wang2021improving} to address disparity, referred to as disparity-aware mutual information regularization (DAMIR), while the second is a novel method focusing on balancing correlation across the dataset, termed as the balanced correlation defense (BCorr).

\subsection{Disparity-Aware Mutual Information Regularization (DAMIR)}
The mutual information regularization method~\cite{wang2021improving} incorporates a secondary loss minimization objective during training to reduce the mutual information between sensitive attribute and output. 
A hyperparameter $\beta$ adjusts the weight of the secondary loss compared to the primary loss, to adjust the strength of the mutual information regularization.
Although it successfully decreases the performance of untargeted attribute inference attacks, it does not mitigate disparity and may even worsen it, as shown in the evaluation results in Figure 7 of ~\cite{dibbo2023model}.
To overcome this limitation, we make a key adjustment: mutual information loss is calculated solely on records from the vulnerable group, ensuring that mutual information between sensitive attribute and output is minimized only for that group’s records.
We refer to this revised defense strategy as disparity-aware mutual information regularization (DAMIR).

\begin{figure}[t]
	\centering
	\begin{minipage}{0.15\textwidth}
	     \includegraphics[width=\linewidth]{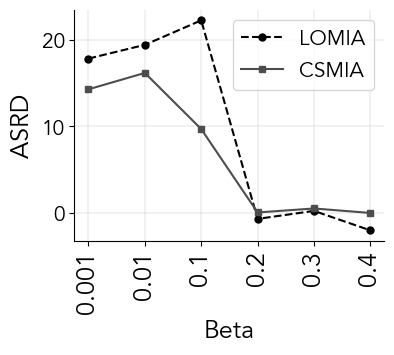}
	      \label{fig:mir_asrd}
	\end{minipage}
	\begin{minipage}{0.15\textwidth}
	    \includegraphics[width=\linewidth]{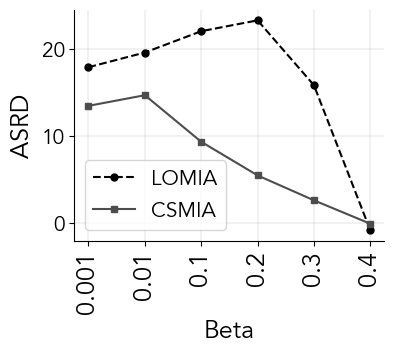}
	    \label{fig:damir_asrd}
        \end{minipage}
	\begin{minipage}{0.15\textwidth}
	    \includegraphics[width=\linewidth]{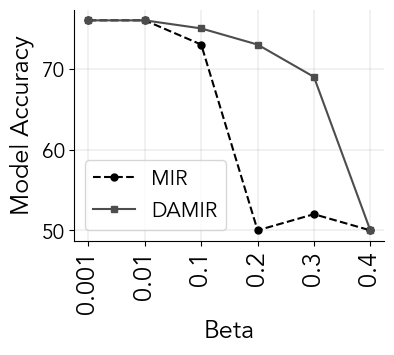}
	\label{fig:mir_damir_model_acc}
         \end{minipage}
\caption{ASRD of Mutual Information Regularization defense (MIR - left, DAMIR - middle) under CSMIA and LOMIA and target model accuracy of MIR and DAMIR trained models (right). }
\label{fig:mutual_info}
\end{figure}

\noindent \textbf{Evaluation}.
We apply both MIR and DAMIR during the training of a subset of the Census-19 dataset where the correlation of Male and Female group is -0.4 and -0.1 respectively.
The range of $\beta$, from 0.001 to 0.4, is used to alter the intensity of the mutual information regularization.
\newtext{
Figure~\ref{fig:mutual_info} shows the attack success rate difference (ASRD) \revisedadd{(defined in section~\ref{sec:bcorr})} between the Male and Female groups and the model accuracy of both MIR and DAMIR trained models.
}
The plots indicate that MIR can only reduce disparity when $\beta$ is set to a high value, resulting in a significant loss of utility in the target model.
DAMIR, in contrast, exhibits slightly better performance in CSMIA, being able to reduce disparity to a degree without substantial degradation of model utility. 
Nevertheless, it can only achieve full disparity mitigation with a considerable loss of model utility.
Furthermore, in LOMIA, the performance is worse, where reducing disparity appears to be impossible without a significant loss of model utility.
Therefore, mutual information regularization is ineffective in reducing disparity.

\subsection{Balanced Correlation Defense (BCorr)}
\label{sec:bcorr}
\revisedadd{
This defense strategy focuses on addressing disparities by ensuring similar correlation levels across all relevant groups in the dataset, thereby eliminating group-specific differences contributing to the disparity.
The main rationale for this defense is that disparity stems from differences in correlation between groups, and thus, mitigating disparity requires eliminating these differences.}
\reviseddelete{
Put simply, this defense strategy involves ensuring a similar correlation level in all relevant groups within the dataset to avoid any disparities.
First, groups in the data that have significantly different correlation levels from the others are identified.
Next, a subset of the data is formed by sampling records from all groups, ensuring that each group’s correlation in the subset aligns with that of the lowest-correlation group.
The ML model is then trained on this curated subset.
Due to space constraints, the formal details of the defense steps are provided in Appendix~\ref{app:balanced_corr_defense}.}

\noindent
\revisedadd{
\textbf{Objective}.
The objective of BCorr is to mitigate disparate vulnerability by lowering attack performance on more vulnerable groups.
In particular, BCorr aims to reduce the metric ASRD (Attack Success Rate Difference) as defined below.
\begin{definition}[ASRD]
\label{def:angular_difference}
    Given a target model $\mathcal{M}$ trained with $\mathbb{D}$ and $\mathbb{D}_1, \mathbb{D}_2, \dots, \mathbb{D}_k$ denoting the set of records belonging to groups defined by a non-sensitive grouping attribute $a$, ASRD is defined as: $ASRD(\mathcal{M}, \mathbb{D}, a, \mathcal{A}) = \;\max_{i \in [1,\;k]}\;{ASR_i} - \min_{j \in [1,\;k]}\;{ASR_j}$ where $\mathcal{A}$ denotes the attack algorithm and $ASR_i = ASR(\mathcal{M}, \mathcal{N}(\mathbb{D}_i), \mathcal{A})$.
\end{definition}}
\noindent
\revisedadd{\textbf{Defense Threat Model}.
To achieve the objective of the defense, we make the following assumptions: the defender has access to the full dataset and the trained target model, and operates as a single entity with complete control over the dataset and model training. Additionally, the defender is aware of which groups are more vulnerable to attacks. This assumption is practical because the defender can simulate attack scenarios using the full dataset and target model to identify vulnerable subgroups. Moreover, the defender can compute correlations between sensitive attributes and outputs for each subgroup, leveraging these correlations as strong indicators of attack vulnerability. These capabilities enable the defender to effectively identify and mitigate disparities in vulnerability.
}

\noindent
\revisedadd{\textbf{Design}.
BCorr comprises of the following steps.
\label{app:balanced_corr_defense}
\begin{itemize}
    \item Given non-sensitive grouping attribute $a$ such that some groups are more vulnerable than others, the first step is to rank groups defined by $a$ in terms of their correlation, ${correlation(\mathcal{S}(\mathbb{D}_i)), \mathbb{Y}(\mathbb{D}_i))}$, where $\mathbb{D}_i \subset \mathbb{D}$ denotes records from group $i$. Suppose, $m$ is the index of the group with the least correlation which is $c_m$.
    \item Sample records from each $\mathbb{D}_i$ to form $\mathbb{D}'_i$ such that the correlation of $\mathbb{D}'_i$ is equal to $c_m$. Note that, $\mathbb{D}_m$ does not need to be sampled and therefore, $\mathbb{D}'_m = \mathbb{D}_m$. Let, $\mathbb{D}' = \mathbb{D}'_{1} \cup \mathbb{D}'_{2} \cup \dots \cup \mathbb{D}'_{k}$. Train model $\mathcal{M}'$ on $\mathbb{D}'$.
\end{itemize}}

\setlength{\tabcolsep}{2.35pt}
\begin{table*}[t]
    \centering
    \tiny
    \begin{tabular}{ccrrrrrrcrrrrr}
         & \multicolumn{13}{c}{Census19} \\
        \hline
         \multirow{3}{*}{Defense} & & \multicolumn{6}{c}{SEX} & & \multicolumn{5}{c}{STATE} \\
         \cline{3-8} \cline{10-14}
          & \phantom{AI} & \multicolumn{2}{c}{ASRD} & \multirow{2}{*}{EOD} & \multirow{2}{*}{DPD} & \multicolumn{2}{c}{MA} & \phantom{AI} & \multicolumn{2}{c}{ASRD} & \multirow{2}{*}{EOD} & \multirow{2}{*}{DPD} & \multirow{2}{*}{MA} \\
         \cline{3-4} \cline{7-8} \cline{10-11} 
         & & CSMIA & LOMIA & & & Male & Female & & CSMIA & LOMIA & & &  \\
        \hline
        None & & 12.52 & 14.97 & 0.0674  & 0.0178  & 73.21  & 78.16 & & 22.94 & 26.75 & 0.4 & 0.2401 & 73.19 \\
        BCorr & & 2.06 & 3.59 & 0.0416  & 0.0015  & 73.89  & 77.90 & & 10.93 & 7.58 & 0.4 & 0.1923 & 74.02 \\
        FC & & 11.45 & 11.90 & 0.0221  & 0.0144  & 70.72  & 70.30 & & {-} & - & - & - & - \\
        \hline
    \end{tabular}
    \hfill
    \begin{tabular}{ccrrrrrrcrrrrr}
         & \multicolumn{13}{c}{Texas-100X} \\
        \hline
         \multirow{3}{*}{Defense} & & \multicolumn{6}{c}{SEX\_CODE} & & \multicolumn{5}{c}{PAT\_STATUS} \\
         \cline{3-8} \cline{10-14}
          & \phantom{AI} & \multicolumn{2}{c}{ASRD} & \multirow{2}{*}{EOD} & \multirow{2}{*}{DPD} & \multicolumn{2}{c}{MA} & \phantom{AI} & \multicolumn{2}{c}{ASRD} & \multirow{2}{*}{EOD} & \multirow{2}{*}{DPD} & \multirow{2}{*}{MA} \\
         \cline{3-4} \cline{7-8} \cline{10-11} 
         & & CSMIA & LOMIA & & & Male & Female & & CSMIA & LOMIA & & &  \\
        \hline
        None & & 12.12 & 14.45 & 0.0332  & 0.0177  & 72.53  & 74.08 & & 17.27 & 20.54 & 0.3707 & 0.2176 & 75.11 \\
        BCorr & & 0.94 & 2.48 & 0.0172  & 0.0186  & 74.70  & 74.35 & & 7.34 & 4.26 & 0.3519 & 0.1982 & 75.19 \\
        FC & & 11.45 & 13.28 & 0.02176  & 0.0020  & 71.788  & 68.108 & & - & - & - & - & - \\
        \hline
    \end{tabular}
    \caption{Comparison of group vulnerability (ASRD), group fairness (EOD, DPD), and group-level model utility (MA) between models trained with and without BCorr Defense.}
    \label{tab:balanced_corr_evaluation}
\end{table*}

\setlength{\tabcolsep}{5pt}
\begin{table}[t]
    \centering
    \tiny
    \begin{tabular}{ccrrrrrr}
        \multicolumn{8}{c}{Census19} \\
        \hline
         & \multirow{2}{*}{MLP Depth} & \multicolumn{2}{c}{ASRD} & \multirow{2}{*}{EOD} & \multirow{2}{*}{DPD} & \multicolumn{2}{c}{MA} \\
         \cline{3-4} \cline{7-8}
          & & CSMIA & LOMIA & & & Male & Female \\
        \hline
         \multirow{3}{*}{BCorr} & 2-layer & 0.45 & 3.05 & 0.0494  & 0.0171  & 75.312  & 78.532 \\
          & 3-layer & 2.06 & 3.59 & 0.0416  & 0.0015  & 73.89  & 77.90 \\
          & 4-layer & 1.45 & 4.54 & 0.0353  & 0.0030  & 72.39  & 75.62 \\
        \hline
    \end{tabular}
    \caption{BCorr performance across varying MLP models.}
    \label{tab:balanced_corr_vs_mlp_architecture}
\end{table}

\noindent
\textbf{Evaluation}.
We examine the effectiveness of the balanced correlation defense with the Census-19 and Texas-100X datasets, specifically targeting the binary attributes \texttt{SEX} and \texttt{SEX\_CODE}, and the multi-valued attributes \texttt{STATE} and \texttt{PAT\_STATUS} for each dataset, respectively.
We also apply a Fairness Constraint-based defense (FC), as used in~\cite{kulynych2022disparate}, to mitigate disparate vulnerability in Membership Inference attacks.
For this baseline, we use the Exponentiated Gradient algorithm with the Equalized Odds~\cite{hardt2016equality} constraint.
The results of this experiment are presented in Table~\ref{tab:balanced_corr_evaluation}. 
We measure group vulnerability by the difference in attack success rates between the most and least vulnerable groups which is referred to as ASRD.
Accuracy is used as the metric for ASRD.
To evaluate group fairness, we use Equalized Odds Difference (EOD)~\cite{cho2020fair} and Demographic Parity Difference (DPD)~\cite{cho2020fair}. Model accuracy (MA) is reported at the group level for binary attribute scenarios and across the entire test dataset for multi-valued attribute scenarios.
BCorr can completely mitigate disparities between Male and Female groups in binary attribute scenarios for both datasets, achieving this without sacrificing model utility or group fairness despite using 66.67\% of the original training data in the balanced correlation set.
FC, in contrast, fails to effectively reduce disparity in either binary attribute scenarios, although it preserves group fairness. 
For multi-valued attribute scenarios, BCorr can substantially reduce disparity and lower the ASR of the most vulnerable group from 73.8\% to 62.92\% for Census19 and from 72.59\% to 59.07\% for Texas-100X under CSMIA.
It is worth noting that reducing ASRD between 51 groups is much harder than between 2 groups, yet BCorr still achieves a substantial reduction.
The FC evaluation for the multi-valued attribute case was omitted because it failed in the binary case, and the computational cost was deemed too high.
BCorr effectively mitigates disparity regardless of MLP depth used in training, as it addresses the root cause of vulnerability at the dataset level, as shown in Table~\ref{tab:balanced_corr_vs_mlp_architecture}.

\noindent
\revisedadd{\textbf{How/Why Bcorr Works}.
Suppose, occupation=nurse and gender=female might constitute the most vulnerable groups within their respective categories in a particular scenario. Female individuals may be more vulnerable than males because a large proportion of them belong to the highly vulnerable ‘nurse’ occupation compared to other occupations like teacher or salesperson. In contrast, there are significantly fewer males in the ‘nurse’ occupation, resulting in a lower overall vulnerability for males.
The assumed scenario is similar to that of the UC Berkeley Admissions Rate Bias study~\cite{bickel1975sex}. Their study found that women faced higher rejection rates overall but were often favored at the department level. This paradox stemmed from women applying disproportionately to competitive departments. In our case, the higher privacy vulnerability of female individuals is driven by their significant representation in ‘nurse’ occupation, a highly vulnerable group.
Through our targeted attack approach, an attacker could determine that ‘nurse’ is the most vulnerable occupation and ‘female’ is the most vulnerable gender.
In our proposed defense, the defender can similarly identify privacy disparities in gender and occupation groups. Depending on the severity of the observed disparities or the defender’s priorities, they can apply BCorr to either attribute (gender/occupation) to mitigate the disparate vulnerabilities effectively. The severity of disparity can be measured through ASRD.
Suppose the disparity across the occupation groups is more severe than that of gender groups and the defender applies BCorr on the occupation attribute. While lowering the attack vulnerability of occupation-based groups, our sampling-based defense approach is unlikely to increase the attack vulnerability of the gender=female group. Specifically, reducing the correlation between the sensitive attribute and model output within the occupation=nurse group—the core mechanism of BCorr—is also likely to reduce the correlation within the female records of that group, resulting in lower attack vulnerability for the gender=female group.}

\noindent
\revisedadd{\textbf{Comparison with Fairness}.
The existing work in fairness~\cite{cho2020fair, cummings2019compatibility, das2024disparate} primarily focuses on ensuring equal model performance across groups. In contrast, BCorr aims to equalize the success rates of attribute inference attacks across groups while simultaneously preserving model performance fairness. Due to this distinction, fairness metrics like equalized opportunity or equalized odds, often used to measure model fairness, are not directly applicable to evaluate BCorr’s ability to mitigate attack disparity. To address this, we introduce ASRD, which parallels fairness metrics like equalized odds difference but measures disparities in attacker success rates across groups rather than disparities in model performance.
Theoretical guarantees for attack disparity mitigation are inherently challenging due to the use of non-linear DNNs and the varying strategies employed by the attacker. Any bound on DNNs must be derived through approximations of their non-linear behavior, which often fail to capture intricate interactions between layers and parameters, resulting in imprecise bounds. Moreover, the variability in attribute inference attack strategies adds another layer of complexity to establishing reliable bounds on attack performance disparity. Nevertheless, we provide empirical evidence that BCorr effectively mitigates disparate vulnerability.
}

\vspace{0.2em}
\section{Related Works}
\vspace{-0.2em}
\textbf{Foundational Works}.
Fredrikson et al. initially introduced model inversion attacks for linear regression models in~\cite{PharmaUSENIX2014} and extended these attacks to non-linear ML models in~\cite{fredrikson2015model}. 
This subsequent work defined two major types of model inversion attacks: attribute inference, in which the adversary tries to uncover the sensitive attributes in the dataset used to train the model, and class representative reconstruction, where the goal is to reconstruct instances resembling those in the training data.
Mehnaz et al. provide the first evidence of disparity in attribute inference attacks~\cite{mehnaz2022your}.

\noindent\textbf{Notable Works in Attribute Inference}.
Existing works on attribute inference attacks have considered a wide range of threat models, with most assuming strong attacker capabilities, rendering their proposed attacks impractical.
In a series of works~\cite{GongTOPS2018, AttriInferWWW2017, GongUSENIX2016}, Gong et al. and Jia et al. applied attribute inference attacks in social media scenarios, where an adversary infers private attributes of a user based on their public information, but these attacks depend on users who also disclose their private attributes publicly, limiting their applicability to cases where private-public attribute pairs can be collected for an auxiliary set of data matching the distribution of the target data.
Yeom et al.~\cite{yeom2018privacy} investigate how the influence, defined as the extent to which changes in a sensitive attribute affect predictions, impacts vulnerability to attribute inference attacks.
Their analysis reveals that while attacker advantage grows with initial increases in influence, it actually decreases as the influence becomes more significant.
Conversely, our research identifies the correlation between the sensitive attribute and the output as a consistent factor contributing to the vulnerability of ML models.
Wang et al.~\cite{wang2021improving} propose a mutual information regularization method to defend against model inversion attacks, including attribute inference attacks.
Mehnaz et al.~\cite{mehnaz2022your} introduce CSMIA and LOMIA, with the latter being the first to show that attribute inference attacks can be launched on models without confidence scores, proving the inadequacy of defenses that mask these scores.
Jayaraman et al.~\cite{jayaraman2022attribute} provide evidence that existing attribute inference attacks perform worse compared to an imputation attack using auxiliary data without even querying the target model.
Our evaluation in section~\ref{sec:uncovering_vulnerability}, however, demonstrates that when there is a high correlation in the private training data, existing attacks surpass the imputation attack in performance.
The study by Tramer et al.~\cite{tramer2022truth} reveals that attribute inference attacks perform better if the adversary can poison the training data. 
However, the assumption that training data can be poisoned is applicable in crowdsourced or collaborative scenarios; the former is unrealistic for private training data, while the latter falls outside the scope of our work.

\noindent\textbf{Other Related Works}.
Dibbo et al.~\cite{dibbo2023model} investigate various potential contributors to disparity but find no consistent factors. 
\revisedadd{While their work focuses on understanding disparity in attribute inference attacks, it does not identify specific contributing factors. Our work goes further by identifying correlation as a key factor driving this disparity.}
\reviseddelete{Their evaluation, unlike ours, does not use datasets with a moderate range of correlation, resulting in an oversight of the strong correlation-attack performance relationship.
~\cite{kulynych2022disparate} suggest using fairness constraints for membership inference attacks, but \cite{dibbo2023model} found them ineffective for attribute inference attacks.}
\revisedadd{Kulynych et al.~\cite{kulynych2022disparate} proposes a defense to mitigate disparate vulnerability in membership inference attacks. In contrast, we tackle attribute inference attacks. We also include [24]’s fairness constraint-based (FC) defense as a baseline in our experiments and demonstrate its ineffectiveness in mitigating disparate vulnerability for attribute inference attacks.}
Zhong et al.~\cite{zhong2023disparate} propose a defense against disparate vulnerability in link inference attacks within GNNs, but their technique cannot be transferred to attribute inference attacks in tabular data.
Property inference attacks, closely related to correlation estimation attacks, were first introduced in~\cite{ateniese2015hacking} and later applied to deep learning models in~\cite{ganju2018property}.
However, existing attacks~\cite{suri2022formalizing, mahloujifar2022property, chaudhari2023snap} focus on inferring properties such as group size rather than the correlation between sensitive attributes and outputs, which is our focus.

\vspace{-0.2em}
\section{Conclusion}
\vspace{-0.2em}
In this paper, we present a series of novel attacks that expose significant privacy leakage in ML models trained on tabular data, a risk that has often been underestimated and overlooked. 
Our comprehensive evaluation reveals that adversaries can identify high-risk groups within records, posing an alarming threat to individuals in those groups. 
Our findings underscore that privacy leakage through targeted attribute inference attacks is far from trivial, with adversaries capable of making highly accurate predictions. 
To address these significant privacy concerns, we introduce a novel defense that have been demonstrated to be effective through evaluation. 
Looking forward, we are committed to incorporating advanced privacy-preserving techniques into our methods to provide a comprehensive toolkit for creating safe and secure data-driven algorithms.






\bibliographystyle{plain}
\bibliography{sample-base}
\appendix
\section{Overview of Notations}
\begin{table}[h!]
    \centering
    \resizebox{0.97\columnwidth}{!}{\begin{tabular}{c|l}
        \hline
        \textbf{Symbol}             & \textbf{Description} \\ \hline
        $\mathbb{D}$                & The original training dataset \\ 
        $\mathcal{M}$               & Target model trained on $\mathbb{D}$ \\
        $Pr({\mathcal{M}}(.))$      & Confidence score output by $\mathcal{M}$ \\
        $\mathcal{A}$               & Attack Algorithm \\
        $s(.)$                      & Sensitive attribute value of a record \\
        $\mathbf{n}(.)$             & Vector of non-sensitive attribute values of a record \\
        $\mathcal{S}$               & Set of possible values of the sensitive attribute \\
        $\mathcal{N}(\mathbb{D})$   & Non-sensitive portion of dataset $\mathbb{D}$ \\
        $\mathcal{C}$               & Confidence matrix \\
        $\rho$                      & Correlation       \\
        \hline
    \end{tabular}}
    \caption{Notations used and their descriptions}
    \label{tab:notation}
\end{table}

\section{Methodology}
\subsection{Correctness Proof of Sampling Technique}
\label{sec:correctness_proof}
Let, $s$ and $y$ denote the sensitive attribute and output respectively and $n$ denote the total number of records.
Then, according to definition,
\begin{equation}
\label{eqn:correlation_def}
    c = \frac{n\sum sy - (\sum s)(\sum y)}{\sqrt{\left( n\sum s^2 - n (\sum s)^2\right)\left( n\sum y^2 - n (\sum y)^2\right)}}
\end{equation}
Now, $\sum y$ essentially means the sum of output values of $n$ records.
Since, the output can either take $0$ or $1$ and takes $1$ a total of $n_+^+ + n_-^+$ times, $\sum y = n_+^+ + n_-^+ $ and $\sum y^2 = n_+^+ \times 1^2 + n_-^+ \times 1^2 = n_+^+ + n_-^+$.
Similarly, $\sum s^2 = \sum s = n_+^+ + n_+^-$.
Substituting these values in equation~\ref{eqn:correlation_def} and performing a few algebraic operations we get,
\begin{equation}
\label{eqn:correlation_eqn_2}
    c = \frac{n_+^+ \times n_-^- - n_+^- \times n_-^+}{\sqrt{(n_+^+ + n_+^-)(n_+^+ + n_-^+)(n_-^+ + n_-^-)(n_+^- + n_-^-)}}
\end{equation}
According to our requirement, the number of positive samples ($n_+^+ + n_-^+$) is equal to the number of negative samples ($n_+^- + n_-^-$) and the ratio between positive and negative samples ($\frac{n_+^+ + n_+^-}{n_-^+ + n_-^-}$) is $m$.
If we combine $n_+^+ + n_+^- + n_-^+ + n_-^- = n$ with the above we get,
\begin{align*}
    n_+^+ + n_-^+ &= n_+^- + n_-^- = \frac{n}{2} \\
    n_+^+ + n_+^- &= \frac{n}{m+1}, \quad n_-^+ + n_-^- = \frac{mn}{m+1}
\end{align*}
Substituting all these in equation~\ref{eqn:correlation_eqn_2} we get,
\begin{equation}
\label{eqn:correlation_eqn_3}
    n_+^+ \times n_-^- - n_+^- \times n_-^+ = c \sqrt{m} \times \frac{n}{m+1} \times \frac{n}{2}
\end{equation}
Substituting $n_+^+$ with $\frac{n}{2} - n_-^+$ and $n_+^-$ with $\frac{n}{2} - n_-^-$ we get,
\begin{equation}
\label{eqn:correlation_eqn_4}
    n_-^- - n_-^+ = c \sqrt{m} \times \frac{n}{m+1} 
\end{equation}
Now we can solve to get the values of $n_-^-$ and $n_-^+$ first, and using those to get the values of $n_+^+$ and $n_+^-$ as shown in section~\ref{sec:experimental_setup}.
Note that, these values are integers which is why the fractions are rounded up using floor and ceiling.
Therefore, the sampled records may not have the exact correlation as $c$ but the difference would be negligible for our purpose.

\reviseddelete{\subsection{Balanced Correlation Defense Process}
\label{app:balanced_corr_defense}
\begin{itemize}
    \item Given non-sensitive grouping attribute $a$ such that some groups are more vulnerable than others, the first step is to rank groups defined by $a$ in terms of their correlation, ${correlation(\mathcal{S}(\mathbb{D}_i)), \mathbb{Y}(\mathbb{D}_i))}$, where $\mathbb{D}_i \subset \mathbb{D}$ denotes records from group $i$. Suppose, $m$ is the index of the group with the least correlation which is $c_m$.
    \item Sample records from each $\mathbb{D}_i$ to form $\mathbb{D}'_i$ such that the correlation of $\mathbb{D}'_i$ is equal to $c_m$. Note that, $\mathbb{D}_m$ does not need to be sampled and therefore, $\mathbb{D}'_m = \mathbb{D}_m$
    \item Let, $\mathbb{D}' = \mathbb{D}'_{1} \cup \mathbb{D}'_{2} \cup \dots \cup \mathbb{D}'_{k}$. Train model $\mathcal{M}'$ on $\mathbb{D}'$.
\end{itemize}}

\section{Details of Experiment Setup}
\subsection{Datasets}
\label{app:datasets}
\noindent\textit{\underline{(1) Census19}}.
Originating from the 2019 US Census Bureau Database~\cite{censusPUMSData}, the Census19 dataset includes over 1.6 million records and 12 variables capturing a wide range of personal and demographic details of US residents.
The goal of this dataset is to classify individuals by their annual income, setting the threshold at over \$90,000, an adjustment from the Adult dataset's \$50,000 threshold to account for inflation over the years.
Marital status is chosen as the sensitive attribute in this dataset. 
The attribute can take multiple values but we convert the attribute into binary by labeling all values except Married as Single.
To streamline analysis, we categorize all instances of marital status into two groups, married or single, in the initial preprocessing steps.

\noindent\textit{\underline{(2) Texas-100X}}.
The Texas-100X dataset expands upon the Texas-100 hospital dataset~\cite{Texas} originally introduced by Shokri et al~\cite{shokri2017membership} and contain 925,128 records from 441 hospitals.
We use the PRINC\_SURG\_PROC\_CODE column from the dataset as the output attribute for this dataset.
However, PRINC\_SURG\_PROC\_CODE is a categorical column that can take up 100 different values.
The other columns do not contain sufficient information related to the surgery procedure to allow training of a target model for a 100-class classification problem with good classification accuracy on a held-out test set.
Therefore, we project the 100 values into 2 distinct categories: top-10 most frequent surgery procedures and the rest of the procedures.
After this mapping, the classification problem becomes a binary one.
For this dataset, SEXCODE is selected as the sensitive attribute which can take up values corresponding to `Male' or `Female'.

\noindent\textit{\underline{(3) Adult}}.
This dataset~\cite{misc_adult_2} is used to predict whether an individual earns over 50,000 a year. 
The dataset contains 48,842 instances and has 14 attributes. 
Following the preprocessing technique in~\cite{mehnaz2022your} We merge the marital status attribute into two distinct clusters: Married, which includes ‘Married-civ-spouse,’ ‘Married-spouse-absent,’ and ‘Married-AF-spouse’; and Single, which includes ‘Divorced,’ ‘Never-married,’ ‘Separated,’ and ‘Widowed.’ 
We then consider this attribute (Married/Single) as the sensitive attribute that the adversary aims to learn. 
After removing records with missing values, the final dataset consists of 45,222 records. 
We split the Adult dataset and use 35,222 records to train the target models, and the remaining 10,000 records to evaluate attacks on data from the same distribution but not in the training set.

\subsection{Model Architecture and Training Hyperparameters}
\label{appendix:model_training}
The default neural network used consists of three hidden layers having 32, 16, and 8 neurons respectively with ReLU activation function. 
For 4-layer MLP, another layer with 64 neuron was added adjacent to input layer and for 2-layer MLP, the layer with 32 neurons were dropped.
The output layer is a softmax layer consisting of one neuron for each output class. 
This is a standard neural network architecture used in prior inference works~\cite{jayaraman2022attribute}.
The Adam optimization algorithm is incorporated with the initial learning rate set to $0.001$.
The training is run for $500$ iterations.

\section{Additional Experiment Results}

\subsection{Model Utility vs. Vulnerability at Group Level}
\begin{figure}
    \centering
    \begin{minipage}{0.23\textwidth}
         \includegraphics[width=\linewidth]{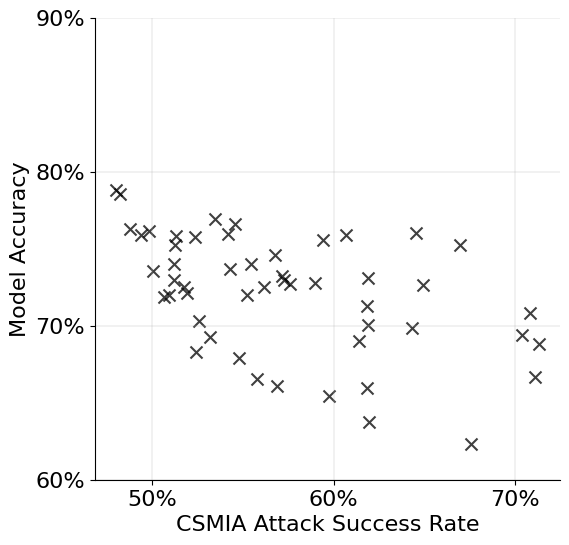}
          \label{fig:utility_vs_vuln_CSMIA}
    \end{minipage}
    \begin{minipage}{0.23\textwidth}
        \includegraphics[width=\linewidth]{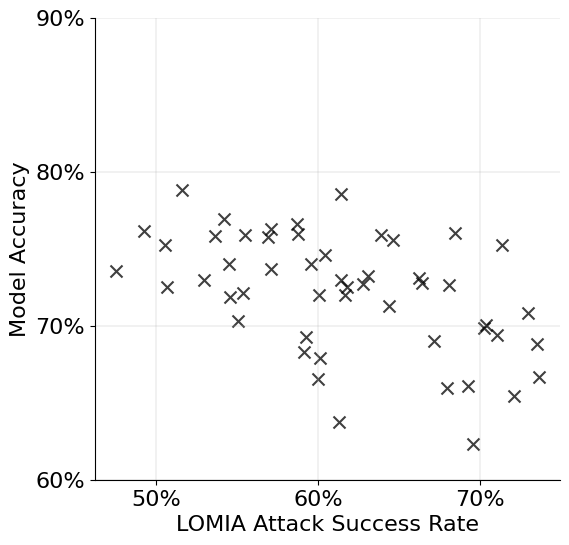}
        \label{fig:utility_vs_vuln_LOMIA}
        \end{minipage} 
\caption{Model utility vs. attack performance (CSMIA - left, LOMIA - right) of 51 states from Census19 dataset. Accuracy is used as a metric for both axes.}
\label{fig:utility_vs_vuln}
\end{figure}

Figure~\ref{fig:utility_vs_vuln} plots the correlation between the sensitive attribute and the output of the 51 states on the X-axis and the angular difference on the Y-axis.
This is from the same scenario as section~\ref{sec:exploiting_disparate_vulnerability}.
The results reveal that there is no correlation between group-level model utility and group vulnerability.
This behavior explains the failure of fairness constraint-based defenses in reducing disparity as groups with similar model utility can have different levels of vulnerability.

\subsection{Correlation vs. Angular Difference}
\begin{figure}[!h]
    \centering
    \includegraphics[width=0.45\textwidth]{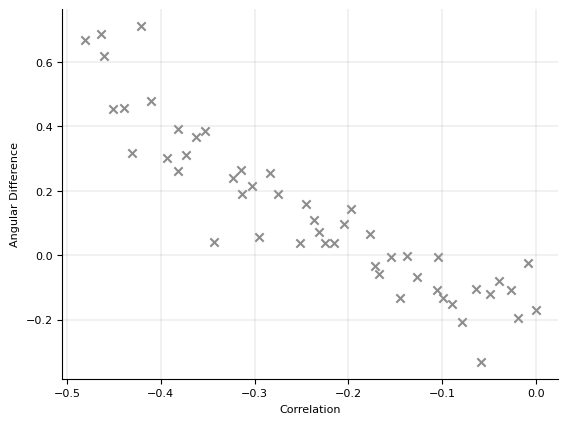}
    \caption{Correlation vs. angular difference of 51 states from Census19 Dataset}
    \label{fig:correlation_vs_angular_difference}
\end{figure}

Figure~\ref{fig:correlation_vs_angular_difference} plots the correlation between the sensitive attribute and the output of the 51 states on the X-axis and the angular difference on the Y-axis.
This is from the same scenario as section~\ref{sec:exploiting_disparate_vulnerability}.
The results reveal that groups with a high correlation between the sensitive attribute and output tend to have a high angular difference, while those with a low correlation exhibit a low angular difference.
While correlation ranges from 0 to -0.5, angular difference ranges from $-0.33$ to $0.71$.
Nonetheless, their relationship is visibly linear confirming the appropriateness of our design choice of fitting a linear regression model for the correlation estimation attack.

\begin{figure}[htbp]
    \centering
    \begin{minipage}{0.4\textwidth}
         \includegraphics[width=\linewidth]{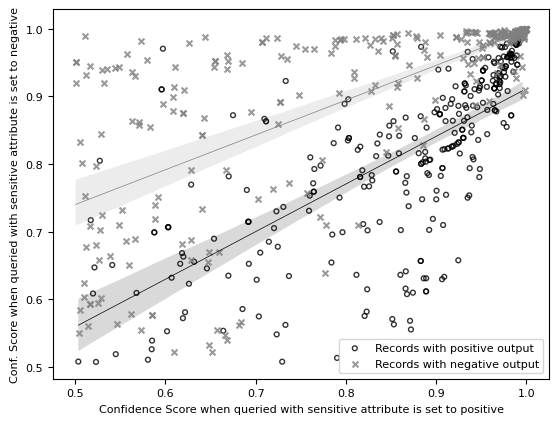}
          \label{fig:confidence_array_state_0}
    \end{minipage}
    \begin{minipage}{0.4\textwidth}
        \includegraphics[width=\linewidth]{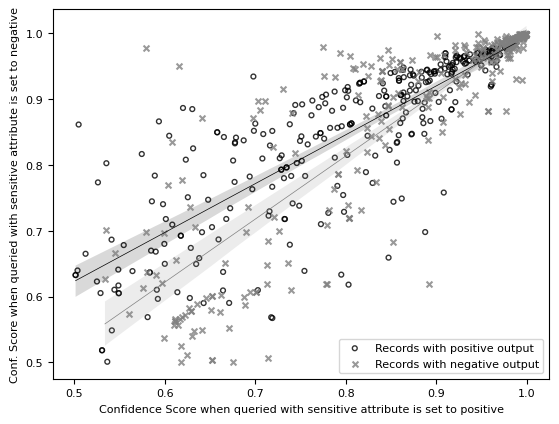}
        \label{fig:confidence_array_state_25}
        \end{minipage} 
    \begin{minipage}{0.4\textwidth}
        \includegraphics[width=\linewidth]{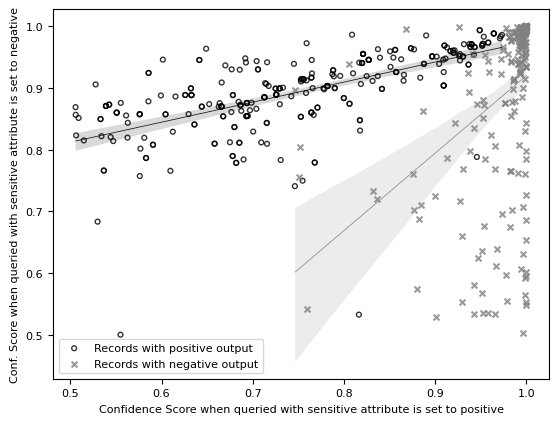}
    \label{fig:confidence_array_state_50}
         \end{minipage}
\caption{Confidence matrix plotted for state with index 0 (left), 25 (middle), and 50 (right). The selected three states have a correlation of 0, -0.25, and -0.5 respectively between the sensitive attribute and output.}
\label{fig:conf_array_plot_for_states}
\end{figure}

\subsection{Angular Difference Visualization Across States}

In Figure~\ref{fig:conf_array_plot_for_states}, we plot the confidence matrix found during the computation of angular difference for the states with indexes 0, 25, and 50 respectively.
We chose these three states particularly as they denote the state with the lowest (0), median (-0.25), and the highest correlation (0.5) among the range of correlation we considered.
In the Census19 dataset, the sensitive attribute is marital status and has two possible values: single and married.
We chose the former as positive and the latter as negative.
The output attribute, which corresponds to the output label given by the target model, also has two values - high income and low income.
High income was chosen as positive and low income was chosen as negative for this attribute.
The positive output records and the negative output records are plotted with different markers and two different regression lines are fit for the two sets of records.
In the highest correlation case, the regression line corresponding to the low income records has a higher slope than the line corresponding to the high income records.
This essentially means that the target model predicted with higher confidence when queried with the negative value for these high income records than when queried with the positive value.
The root of this behavior comes from the high magnitude of correlation we set for this particular group as we identified in section~\ref{sec:uncovering_vulnerability}.
A high negative correlation means that there are more married high income records than their single counterpart.
In summary, the plots reveal that the variation in angular difference across groups arises inherently from the variation in the correlation between the sensitive attribute and output.

\clearpage

\end{document}